\documentclass[]{fairmeta}

\usepackage[utf8]{inputenc}
\usepackage{amsfonts}
\usepackage{amsmath}
\usepackage{amssymb}
\usepackage{mathtools}
\usepackage{amsthm}
\usepackage{nicefrac}
\usepackage{float}
\usepackage{enumitem}
\usepackage{algorithm}
\usepackage{algorithmic}
\usepackage{tikz}
\usetikzlibrary{positioning}
\usepackage{wrapfig}

\renewenvironment{wraptable}[2]{\begin{table}[htbp]}{\end{table}}

\newcommand{\ourmodel}{FlowWM}
\newcommand{\ourtask}{\textsc{FuturePerception}}

\title{Flow Matching in Feature Space for Stochastic World Modeling}

\author[1]{Fran\c{c}ois Porcher}
\author[1]{Nicolas Carion}
\author[2]{Karteek Alahari}
\author[2]{Shizhe Chen}

\affiliation[1]{FAIR at Meta}
\affiliation[2]{Inria}

\date{\today}
\correspondence{Fran\c{c}ois Porcher at \email{francoisporcher@meta.com}}

\metadata[Code]{\url{https://github.com/facebookresearch/Flow-World-Models}}

\abstract{%
World modeling requires forecasting uncertain futures while preserving information useful for downstream perception.
Existing visual world models often struggle to satisfy both goals: VAE-based stochastic models operate in low-dimensional reconstruction latents, which can limit perception performance, while deterministic predictors using strong pretrained features collapse multimodal futures into a single blurry mean.
In this work, we propose \emph{\ourmodel}, a stochastic world model that performs flow matching directly within pretrained feature space (e.g., DINOv3).
This is challenging because pretrained features are substantially high-dimensional, making standard diffusion recipes suboptimal.
To address this, we investigate the design choices needed for feature-space flow matching and introduce a differentiable one-step projection mechanism that enables efficient training with temporal consistency and task-driven objectives.
We evaluate \ourmodel{} on two benchmarks: a synthetic benchmark for systematic evaluation of accuracy and diversity, and a real-world benchmark \ourtask{}.
\ourmodel{} improves perception performance, mode coverage, and horizon robustness, validating our proposed design for stochastic world modeling in high-dimensional feature spaces.%
}

\begin{document}

\maketitle

\section{Introduction}

World models aim to forecast future states from past context~\citep{ha2018worldmodels}.
In visual domains like autonomous driving~\citep{vasudevan2025planning} and robotics~\citep{wu2023daydreamer}, effective forecasting requires more than generating visually plausible pixels: the predicted states should preserve information relevant to downstream perception and planning~\citep{maes2026leworldmodel}. This makes the choice of representation central to visual world modeling.

Recent advances in video generative modeling~\citep{liu2024sorareviewbackgroundtechnology, peng2025opensora20trainingcommerciallevel, wan2025wanopenadvancedlargescale} have driven diffusion~\citep{ho2020denoising, song2021score} and flow-matching models~\citep{lipman2023flow, liu2022rectified} as visual world models.
These methods first compress videos into a low-dimensional latent space using a VAE~\citep{alonso2024diffusionworldmodelingvisual}, then train neural networks~\citep{peebles2023dit,ronneberger2015unet} to model the latent dynamics. 
However, VAE latents are optimized for pixel-wise reconstruction rather than geometry or semantics, and thus are inadequate for perception tasks that are critical for high-level planning such as object localization, detection, and tracking \citep{kumar2022sharpness, jayasumana2024rethinking, yao2025reconstruction, shi2025latentdiffusionmodelvariational}.

\begin{figure}[t]
\centering
\includegraphics[width=0.9\linewidth]{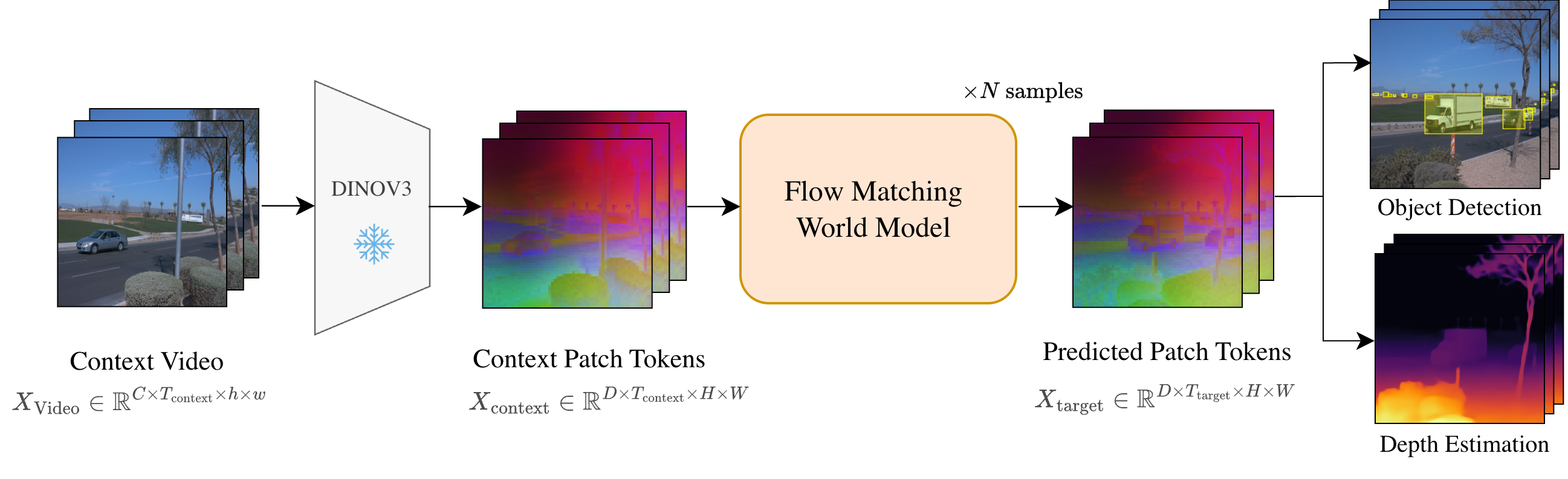}
\caption{
\textbf{Overview of \ourmodel}.
A frozen DINOv3 backbone encodes context video frames to patch tokens, and our Flow Matching World Model samples possible futures in the same latent space. The predictions are evaluated on perception tasks such as object detection and depth estimation.}
\label{fig:intro_overview}
\vspace{-1.1em}
\end{figure}

An alternative is to predict future states in the feature space of pretrained visual encoders such as DINOv3~\citep{simEoni2025dinov3} and VJEPA-2 \citep{assran2025v}.
These representations excel at capturing semantic correspondence and objectness, showing significant promise for perception-oriented future prediction~\citep{zhou2024dinowm,karypidis2024dinof}.
However, current feature-space world models remain largely deterministic, yielding a single future for the given context.
This is a critical limitation in complex real-world environments like autonomous driving, where the future is inherently multimodal and stochastic: a vehicle may brake or continue, a pedestrian may cross or stop, and occluded objects may appear or remain hidden. 
Deterministic predictors trained with standard regression losses tend to average over these possibilities, producing predictions that may correspond to no valid future (see formal proof in Appendix~\ref{sec:appendix_stochastic_necessity}).

To overcome this limitation, recent work~\citep{walker2025generalist} has explored stochastic world modeling by applying a latent diffusion model to frozen pretrained video features. 
Yet, adopting the standard DiT recipe~\citep{peebles2023dit} for low-dimensional VAE latents is unstable or suboptimal in a significantly higher-dimensional space~\citep{mousakhan2025orbis}. 
Similar challenges have been observed in image generation, where diffusion in pretrained representation spaces requires designs tailored to high-dimensional latents~\citep{zheng2025diffusiontransformersrepresentationautoencoders}. 
Video forecasting further compounds these difficulties, since generated feature sequences must remain temporally coherent across future frames.

To address these issues, we introduce~\ourmodel{}, a stochastic world model that leverages flow matching in high-dimensional feature spaces (e.g., DINOv3~\citep{simEoni2025dinov3}).
We identify and propose the key components needed for feature-space generative modeling, including architectural design, diffusion timestep scheduling, and training objectives for temporal coherence and task relevance.
To apply these objectives efficiently, we introduce a differentiable one-step projection mechanism that supervises predicted future features without backpropagating through the full sampling trajectory.

We construct two complementary benchmarks for systematic evaluation: a synthetic stochastic bouncing-shapes environment and a real-world benchmark~\ourtask{} derived from the Waymo Open Dataset~\citep{Sun_2020_CVPR}.
The synthetic benchmark provides enumerated multimodal futures, enabling controlled evaluation of prediction accuracy, diversity, and mode coverage.
The real-world benchmark evaluates predicted future features through essential downstream perception tasks, including object detection and depth prediction.
Across these benchmarks,~\ourmodel{} improves perception performance, mode coverage, and horizon robustness compared with deterministic predictors~\citep{zhou2024dinowm} and stochastic models operating in either VAE~\citep{wan2025wanopenadvancedlargescale} or pretrained feature spaces~\citep{walker2025generalist}.
These results demonstrate the effectiveness of the proposed method for stochastic world modeling in high-dimensional feature space.

To summarize, our contributions are as follows:

\begin{itemize}[itemsep=0.1em,parsep=0em,topsep=0em,partopsep=0em]
    \item We propose~\ourmodel{}, a stochastic world model operating directly in pretrained high-dimensional feature spaces.
    \item We investigate design choices required for stochastic feature-space world models with flow matching, including architecture, timestep scheduling, and efficient temporal consistency and task-driven training via one-step projection.
    \item We introduce synthetic and real-world benchmarks for evaluation, showing that~\ourmodel{} produces state-of-the-art perception performance, mode coverage, and horizon robustness.
\end{itemize}
We will release the code and benchmarks to accelerate research in this direction.

\vspace{-.5em}
\section{Related Work}
\vspace{-.5em}

\noindent\textbf{World Modeling.}
World models learn to predict future environment states, enabling imagined rollouts for planning and decision making~\citep{ha2018worldmodels}.
Classical approaches often leverage VAE-based latents~\citep{van2017vqvae} within action-conditioned dynamic models~\citep{hafner2019dream,hafner2019learning,micheli2022transformers,kaiser2019model,bruce2024genie,gao2025adaworld}.
While effective in quasi-deterministic settings, they often suffer from representation collapse, making them unsuitable for more stochastic environments~\citep{mousakhan2025orbis}.
World models such as DINO-WM~\citep{zhou2024dinowm} and DINO-Foresight~\citep{karypidis2024dinof} instead predict future states in pretrained visual features~\citep{oquab2023dinov2}, demonstrating the benefits of semantically rich representations for robotics and perception.
Despite these advances, these models remain largely deterministic~\citep{hansen2023td,zhou2024dinowm,karypidis2024dinof,assran2025v}, limiting their ability to represent uncertainty and multi-modality, particularly over long horizons.
Modeling stochastic dynamics in high-dimensional spaces remains challenging~\citep{walker2025generalist,mousakhan2025orbis}.
Orbis~\citep{mousakhan2025orbis} avoids this difficulty by distilling DINO features into lower-dimensional VAE latents.
The work most closely related to ours is~\citet{walker2025generalist}, which applies diffusion directly to frozen visual features for stochastic forecasting.
However, its main focus is on comparing frozen representations rather than on improving stochastic world models.
In contrast, our work improves stochastic world modeling in high-dimensional latent spaces.

\noindent\textbf{Video Generative Modeling.}
Video generative modeling learns a video distribution $p(x_{1:T})$, typically with flow matching or diffusion models~\citep{liu2024sorareviewbackgroundtechnology,peng2025opensora20trainingcommerciallevel,wan2025wanopenadvancedlargescale}.
Most methods adopt a two-stage pipeline in which a spatiotemporal VAE compresses videos into a low-dimensional latent space, followed by generative modeling in that space~\citep{deng2025autoregressivevideogenerationvector}. While this yields high visual fidelity, strong VAE bottlenecks discard fine-grained semantic structure, making the resulting representations poorly suited for dense perception and planning tasks~\citep{karypidis2024dinof, Jenni_2018_CVPR, newell2020useful}.
To improve diffusion model training,  recent works incorporate task-driven objectives by backpropagating differentiable rewards through the diffusion inference process~\citep{xu2023imagerewardlearningevaluatinghuman,clark2024directlyfinetuningdiffusionmodels}. This requires differentiating through the full or truncated denoising trajectory, which is computationally expensive and prone to vanishing or exploding gradients. 
Instead, we use a one-step projection that applies downstream supervision at the endpoint, avoiding backpropagation through time while remaining stable and efficient.

\noindent\textbf{Diffusion in High-Dimensional Latent Space.}
%
Recent work has shown that diffusion models can operate directly in high-dimensional latent spaces. RAE~\citep{zheng2025diffusiontransformersrepresentationautoencoders} achieves state-of-the-art image generation by performing diffusion in the latent space of DINO, using a large projection head and a carefully designed time-sampling strategy. Similarly, Just Image Transformers (JIT)~\citep{li2025basicsletdenoisinggenerative} demonstrates that diffusion transformers scale to high-dimensional spaces by projecting directly onto the data manifold rather than predicting a velocity field.
While these works focus on image generation, we study stochastic world models, where the model must not only generate plausible future features but also maintain temporal coherence over long prediction horizons.

\section{The Proposed Method: \ourmodel{}}
\vspace{-.5em}

\subsection{Problem Formulation: Latent Space World Modeling}

We aim to train world models \emph{directly in the high-dimensional latent spaces} of pretrained vision encoders, leveraging their strong semantic representations for downstream perception.

Given a video sequence $I_{1:T}$ of raw frames, the first $T_{\text{context}}$ frames are encoded frame-by-frame using a frozen pretrained encoder $E$ (e.g., DINOv3~\citep{simEoni2025dinov3}), and the model must predict the subsequent latent frames of length $T_{\text{target}}$.
Formally, we define the encoded context latents, the feature-space representations of observed frames, as
    $x_{\mathrm{ctx}} = x_{1:T_{\text{context}}} = E(I_{1:T_{\text{context}}})$,
where $x_{\mathrm{ctx}} \in \mathbb{R}^{T_{\text{context}} \times H \times W \times D}$ with $H, W, D$ denoting the spatial height, width and channel dimension of the latent feature map, respectively.
The objective is to generate future latents
    $\hat{x}_{\mathrm{future}} = \hat{x}_{T_{\text{context}}+1:T} \in \mathbb{R}^{T_{\text{target}} \times H \times W \times D}$.

Evaluation is performed on downstream \emph{perception tasks} using pretrained, frozen decoders applied to the predicted features.
This evaluation explicitly probes the preservation of object-centric information and scene geometry which are essential for planning and control, whereas low-level reconstruction metrics that capture pixel- or feature-level statistics may not correlate with task utility.

\subsection{Stochastic World Model with Flow Matching}

To address the limitations of deterministic latent world models, we introduce~\ourmodel{}, a stochastic latent space world model with flow matching, as illustrated in Figure~\ref{fig:intro_overview}.

\noindent\textbf{Flow Matching.}
We use flow matching~\citep{lipman2023flow,liu2022rectified} with the standard linear probability path.
Let $x_0$ denote a noise latent and $x_1$ a data latent corresponding to future frames conditioned on a context latent $x_{\mathrm{ctx}}$.
We use $\tau \in (0,1)$ to denote the continuous interpolation variable along the probability path\footnote{We reserve $t$ exclusively for the model's input time coordinate to avoid confusion with $\tau$.}, and a linear interpolation between noise and data is computed as:
\begin{equation}
    x_\tau = (1 - \tau)\, x_0 + \tau\, x_1, \qquad \tau \in [0,1].
\end{equation}
For this linear path, the ground-truth velocity field is
    $u^\star(x_\tau, \tau) = x_1 - x_0$.
%
A neural network $u_\theta(x_\tau, x_{\mathrm{ctx}}, \tau)$ is trained to approximate this velocity, thereby learning the conditional distribution $p(x_{\text{future}} | x_{\mathrm{ctx}})$.
Given a trained velocity field, sampling is performed by integrating the ODE
$\frac{d x_\tau}{d \tau} = u_\theta(x_\tau, x_{\mathrm{ctx}}, \tau),$
starting from $x_0 \sim p_{\text{noise}}$ and integrating from $\tau = 0$ to $\tau = 1$ to obtain a final prediction $\hat{x}_1$.

\begin{wrapfigure}{r}{0.36\linewidth}
\vspace{-2.em}
  \centering  \includegraphics[width=1\linewidth]{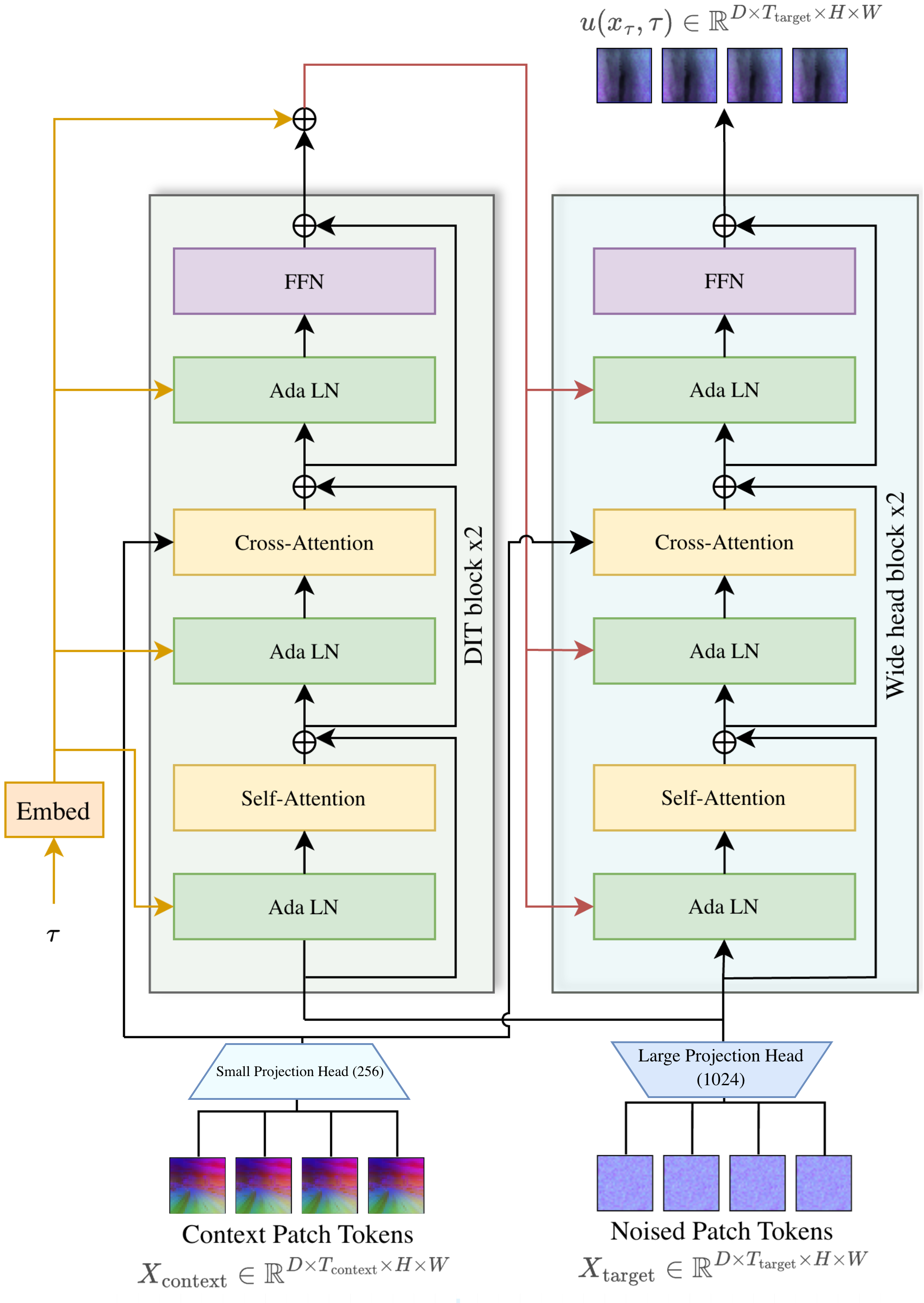}
  \caption{\textbf{Overview of our~\ourmodel{} architecture.} The model takes as input a noisy target latent $x_\tau$, conditioned on encoded context frames $x_{\mathrm{ctx}}$, to predict the flow field $u_\theta(x_\tau, \tau)$.
  }
  \label{fig:architecture}
  \vspace{-3em}
\end{wrapfigure}

\noindent\textbf{Model Architecture.}
We adopt a Transformer-based latent generative architecture operating directly in pretrained semantic representations (Figure~\ref{fig:architecture}).
The model takes a noisy target latent $x_\tau$ as input and is conditioned on encoded context frames $x_{\mathrm{ctx}}$ via cross-attention.
We decompose positional embeddings into temporal and spatial components and apply rotary positional encodings (RoPE)~\citep{su2024roformer} to both $x_\tau$ and $x_{\mathrm{ctx}}$.

The model consists of a backbone for multimodal fusion and a projection head for velocity prediction.
The backbone contains $2$ DiT-style Transformer blocks with dimension $256$, using query--key normalization~\citep{henry2020query} for stability.
The resulting features are fed through AdaLN conditioning into a shallow but wide projection head~\citep{zheng2025diffusiontransformersrepresentationautoencoders} composed of $2$ wide projection layers (projection dimension $1024$), which predicts the flow field $u_\theta(x_\tau,\tau)$.
The wide projection dimension ($1024$) ensures the head's hidden dimension exceeds the per-patch latent dimension ($d{=}384$), which is critical for accurately predicting velocity fields in high-dimensional latent spaces.

\subsection{Training Objectives}

\noindent\textbf{Standard Flow Matching Objective.}
We optimize the model using the conditional expected squared error~\citep{lipman2023flow}:
\begin{equation}
\mathcal{L}_{\text{FM}}(\theta)
= \mathbb{E}_{\substack{x_1\sim p_{\text{data}},\,x_0\sim p_{\text{noise}},\\ \tau\sim\mathcal{U}(0,1)}}\!\left[\left\|u_\theta(x_\tau,\tau)-(x_1-x_0)\right\|_2^2\right].
\end{equation}

\noindent\textbf{Time Consistency with One-Step Projection.}
\label{sec:time-consistency}
Video generation requires \emph{temporal consistency}, ensuring that dynamic entities adhere to plausible motion trajectories.
Achieving such consistency remains a central challenge, especially as the temporal horizon increases, with existing models exhibiting artifacts like flickering or inconsistent motions~\citep{ho2022videodiffusionmodels,yan2023temporallyconsistenttransformersvideo}.

To address this, we propose a regularization term inspired by Physics-Informed Neural Networks~\citep{raissi2019pinns} that constrains not only the model output, but also its derivatives.
Specifically, we regularize predicted latents to respect the time derivative of ground-truth latents.

Directly imposing losses on the generated endpoint $x_1$ requires integrating the ODE, which is computationally expensive and prone to vanishing or exploding gradients~\citep{hu2021loralowrankadaptationlarge,xu2023imagerewardlearningevaluatinghuman}.
Instead, we propose a \emph{one-step projection} to obtain a differentiable estimate of the predicted endpoint $x_1$ without integrating the full flow.

\begin{wrapfigure}{l}{0.38\linewidth}
  \centering
  \includegraphics[width=1\linewidth]{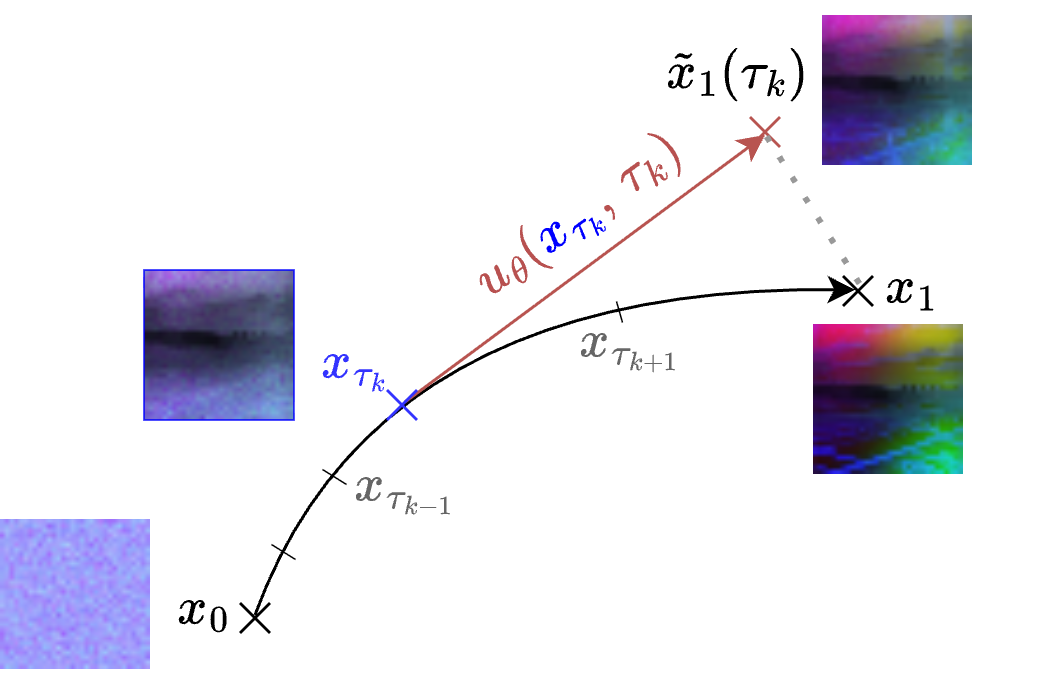}
  \caption{\textbf{One-step projection in Flow Matching.} A differentiable estimate of the final endpoint is obtained from $x_\tau$ by adding the predicted velocity, yielding $\tilde{x}_1(\tau)$ without integrating the full flow.}
  \label{fig:one_step_projection}
\end{wrapfigure}

Given $x_{\tau}$ at time $\tau$, we approximate the denoised endpoint via a linear projection along the current velocity vector $u_\theta$:
\definecolor{diagramred}{RGB}{200,40,40}
\begin{equation}
\tilde{x}_1(\tau)
= \textcolor{blue}{x_\tau} + (1-\tau)\,\textcolor{diagramred}{u_\theta({\textcolor{blue}{x_\tau}},\tau)}.
\label{eq:one_step_projection}
\end{equation}


Using this one-step projection, we impose temporal consistency directly on the projected endpoint.
Let $x_t^{\text{GT}}$ denote the ground-truth latents of the frame at time $t$.
We define the ground-truth temporal derivative as:
\begin{equation}
    \Delta x_t^{\text{GT}} = x_{t+1}^{\text{GT}} - x_t^{\text{GT}}.
\end{equation}
We then enforce alignment between the ground-truth temporal differences and those of the predicted latents obtained via one step projection Eq.~(\ref{eq:one_step_projection}), yielding an auxiliary loss:
\begin{equation}
    \mathcal{L}_{\text{temporal}} = \sum_{t=1}^{T-1} \left\| \Delta \tilde{x}_t - \Delta x_t^{\text{GT}} \right\|^2,
\label{eqn:temporal_consistency_loss}
\end{equation}
where $\tilde{x}_{1:T}$ are the projected endpoints of video frames.
This encourages the predicted latents at the end of the flow to have consistent temporal derivatives as real latents.

\noindent\textbf{Task-driven Objective.}
\label{sec:detector-backprop}
The flow matching and temporal consistency objectives enable training general latent world models without any task-specific supervision.
A natural question then arises:
\emph{if the downstream task is known at training time, can it provide additional supervision to further shape the world model?}

The one-step projection introduced in Eq.~(\ref{eq:one_step_projection}) provides an efficient and stable mechanism to incorporate task-driven objectives (e.g., manifold regularization, differentiable rewards, or detector-based losses) into flow-matching training.

Assume a downstream perception task such as object detection with a pretrained object detector operating in the latent space.
We first leverage the one-step projection to obtain a differentiable estimate of the final predicted latent state, denoted by $\tilde{x}_1$.
The frozen detector is then applied to this projected endpoint, and its detection loss $\mathcal{L}_{\text{det}}(\tilde{x}_1)$ is backpropagated only through the projection and the world model.
In this way, downstream task supervision actively steers the world model towards representations that are more semantically useful for the downstream application.

\subsection{Shifting of Timestep Schedules}
\label{sec: timestep_shifting}
When increasing the dimension of the tensor being generated, both the timestep sampling distribution during training and the inference scheduler must be adjusted to maintain a stable signal-to-noise ratio.

We use the resolution-dependent timestep shift of~\citet{esser2024scalingrectifiedflowtransformers}, which rescales diffusion time by a factor $\alpha$, toward noisier timesteps. Although this strategy was initially introduced for generating images in higher resolution, the same principle applies when scaling latent dimension or prediction horizon, since both increase the number of noise dimensions.
Specifically, the original timestep $\tau \in [0,1]$ is transformed as
\begin{equation}
\tau' \;=\; \frac{\alpha\, \tau}{1 + (\alpha - 1)\, \tau}\,,
\end{equation}
which preserves the endpoints while redistributing probability mass toward larger noise levels.

\section{Latent World Model Benchmarks}
\label{sec:benchmarks}

To systematically benchmark latent world models, particularly focusing on perception evaluation which is more relevant for planning and control than pixel-wise generation, we introduce two complementary benchmarks spanning synthetic and real-world environments.
The synthetic benchmark provides controlled stochastic settings with simple objects, enabling comprehensive and rigorous evaluation of model quality.
In contrast, our \ourtask{} benchmark presents richer visual complexity and dynamics, but offers less rigid evaluation due to limited ground-truth structure.

\subsection{Synthetic Benchmark: Bouncing Shapes}
\label{sec:bouncing_shapes_benchmark}
\vspace{-0.25em}

The \emph{Bouncing Shapes} benchmark defines a controlled stochastic environment in which two objects (a red square and a blue ball) move inside a 2D box and bounce off the walls.
At each wall collision, an object either performs a standard bounce or reverses its incoming velocity with probability $0.5$, inducing an exponential branching of possible futures.
We can enumerate \emph{all} futures consistent with the observed context.
We use $16$ context frames to predict the next $16$ frames.

\noindent\textbf{Evaluation metrics.}
We evaluate the future latent prediction quality via a pretrained decoder that predicts the 2D coordinates of the objects from the latent space.
Since all possible futures can be enumerated, we can measure metrics that are not available in real datasets:
\\
$\bullet$ \emph{Precision Error:} Do sampled predictions correspond to valid modes? Each prediction is matched to its nearest ground-truth mode, and the average matching error over all predictions is reported.
\\
$\bullet$ \emph{Recall Error:} Does the model cover all valid modes? Each ground-truth mode is matched to its nearest sampled prediction, and the average matching error over all ground-truth modes is reported.

We also define an F1 error as the harmonic mean of Precision Error and Recall Error.
The details of the benchmark and metrics are provided in Appendix~\ref{sec:appendix_bouncing_shapes_benchmark}.

\vspace{-0.5em}

\subsection{Real-world Benchmark:~\ourtask{}}
\label{sec:waymo_benchmark}
\vspace{-0.25em}

We construct our real-world benchmark based on the Waymo Open Dataset~\citep{Sun_2020_CVPR}, originally introduced for object tracking.
Although \citet{walker2025generalist} proposed a similar protocol, no code or splits were released, so we recreate our~\ourtask{} task on top of Waymo.
No data filtering is applied, so the dataset preserves the natural distribution of real-world autonomous driving scenarios.
The task uses 4 context frames to predict the next 12 frames, which corresponds to 1.2 seconds into the future at 10 FPS.
Examples are provided in Appendix~\ref{sec:appendix_waymo_benchmark}.

The~\ourtask{} task is challenging for three main reasons:
i) \emph{Resolution.} Waymo videos are high-resolution ($1920 \times 1280$), making diffusion- and flow-based video modeling substantially more difficult than on ImageNet-scale datasets ($256 \times 256$).
ii) \emph{Dynamics.} Real-world driving scenes involve many interacting agents with complex motion patterns, including occlusions and viewpoint changes induced by ego-motion (e.g., vehicles passing behind others or pedestrians temporarily disappearing).
iii) \emph{Stochasticity and Partial Observability.} The future is inherently multi-modal and partially observed. Given the same context, agents may take different plausible actions (e.g., braking or continuing), while new objects may enter the scene or previously occluded ones may reappear, requiring the model to represent uncertainty rather than commit to a single deterministic outcome.

\noindent\textbf{Evaluation metrics.}
The latent prediction quality is evaluated in  downstream object detection and depth prediction tasks.
For object detection, we focus on three classes: vehicle, pedestrian, and cyclist, and restrict the evaluation to large objects, for which detection is reliable at this resolution.

We evaluate detection performance using \textbf{AP$_{\text{L}}$(N)}, which reports the maximum AP$_{\text{L}}$ across $N$ sampled predictions.
This best-of-$N$ metric captures the model's capacity to represent multi-modal futures, rather than averaging over incompatible outcomes.

For depth prediction, 
we report standard monocular depth metrics~\citep{eigen2014depth}: RMSE and threshold accuracy $\delta_i$ (fraction of pixels where $\max(d/d^*, d^*/d) < 1.25^i$, for $i \in \{1,2,3\}$).

\vspace{-0.5em}
\section{Experiments}
\vspace{-0.5em}

\subsection{Experimental Setups}
\vspace{-0.25em}

\noindent\textbf{Baselines.}
We compare~\ourmodel{} against the following three classes of baselines:
\parskip=0.1em
\begin{itemize}[itemsep=0.1em,parsep=0em,topsep=0em,partopsep=0em,leftmargin=1.em]
    \item Deterministic world models using high-dimensional latents: we consider DINO-WM~\citep{zhou2024dinowm} which predicts future latents \emph{autoregressively}, and a stronger deterministic baseline that predicts the entire future latent trajectory \emph{jointly}, using the same architecture as~\ourmodel{} for a fairer comparison. The DINOv3 encoder ($D=384$) is used for both models. 
    \item Stochastic world models using low-dimensional VAE latents: we replace the DINOv3 encoder with the VAE encoder used in WAN~2.2~\citep{wan2025wanopenadvancedlargescale}, which compresses each frame into a low-dimensional latent representation ($D=16$).
    \item State-of-the-art stochastic world models using high-dimensional latents: we re-implement the model~\citep{walker2025generalist} with DINOv3 features following discussions with the authors.
\end{itemize}

\noindent\textbf{Implementation Details.}
We use a frozen DINOv3 ViT-S model~\citep{simEoni2025dinov3} as encoder $E$ and encode each frame separately. We average pool token features from every third layer (3, 6, 9, 12), motivated by findings that intermediate layers can provide stronger representations~\citep{bolya2025perceptionencoderbestvisual}.
We train all models using AdamW with a learning rate of $10^{-3}$ and an effective batch size of 128 for 52K gradient steps (150 epochs total).
We apply gradient clipping with a 1.0 norm, use a linear warmup for 10 epochs, and cosine annealing afterwards.
Images from the Waymo dataset are resized to $512 \times 512$, and are augmented with random horizontal flip.
The model is trained with 64 NVIDIA V100 GPUs in fp32 precision, taking roughly 25 hours.
For inference, we use the Euler solver with sampling over 50 steps. 
For object detectors, we use the DINO-DETR architecture~\citep{zhang2022dinodetrimproveddenoising} based on DETR~\citep{carion2020endtoendobjectdetectiontransformers}, implemented in the detrex codebase~\citep{ren2023detrex} and relying on detectron2~\citep{wu2019detectron2}.
For depth prediction, we train a lightweight depth head on DINOv3/WAN VAE latents using pseudo-labels from Depth Anything V2~\citep{yang2024depth}.

\begin{wraptable}{r}{0.5\linewidth}
\caption{
\textbf{Results on the Bouncing Shapes benchmark.} TC denotes temporal consistency loss.
We report precision, recall, and F1 errors, where lower is better. The Oracle uses ground-truth future latents. Lower precision error indicates more accurate samples; lower recall error indicates greater diversity and mode coverage. 
}

  \label{tab:sbs_errors}
  \begin{center}
    \setlength{\tabcolsep}{4pt}
    \begin{small}
      \renewcommand{\arraystretch}{1.2}
      \begin{tabular}{@{}lrrr@{}}
        \toprule
        Type & Prec.\ $\downarrow$ & Rec.\ $\downarrow$ & F1 $\downarrow$ \\
        \midrule
        Dino-WM & 16.1 & 19.9 & 17.8 \\
        Deterministic Predictor & 15.7 & 18.8 & 17.1 \\
        \cite{walker2025generalist} & 14.5 & 14.3 & 14.4 \\
        \ourmodel{} w/o TC & 4.57 & 4.49 & 4.53 \\
        \textbf{\ourmodel{} (Ours)} & \textbf{4.35} & \textbf{4.28} & \textbf{4.31} \\
        \midrule
        Oracle & \textit{1.01} & \textit{0.98} & \textit{1.00} \\
        \bottomrule
      \end{tabular}
    \end{small}
  \end{center}
\end{wraptable}

\subsection{Results on the Synthetic Benchmark}
\label{sec:bouncing_shapes_results}

Table~\ref{tab:sbs_errors} reports performance of deterministic and stochastic world models using the same high-dimensional DINOv3 features on our synthetic benchmark.
As the task is relatively simple, we do not apply the task-driven objective to train our model.
For reference, we report the Oracle performance of evaluating the coordinate predictor on ground-truth future latents.

\ourmodel{} consistently outperforms deterministic predictors, including DINO-WM and our deterministic variant with the same architecture, across all evaluation metrics. This confirms the importance of explicitly modeling stochasticity in multimodal future prediction.
The stochastic baseline of \citet{walker2025generalist} improves over deterministic models, but remains substantially worse than \ourmodel{}, suggesting that standard flow-matching designs are suboptimal in high-dimensional feature spaces.
In contrast, \ourmodel{} sharply reduces the F1 error from 14.4 to 4.53 even without temporal consistency. Adding the temporal consistency loss further improves the F1 error to 4.31.
These gains come from improvements in both precision and recall: sampled trajectories are closer to valid futures, while the model also covers the set of plausible futures more effectively.

\subsection{Results on the Real-World Benchmark}
\label{sec:waymo_results}

We further evaluate scalability of our approach on the more challenging and realistic \ourtask{} benchmark on downstream object detection and depth estimation tasks.

\noindent\textbf{Object detection.}
As shown in Table~\ref{tab:future_detections_joint}, \ourmodel{} substantially outperforms both deterministic and VAE-based world models.
The gap to deterministic baselines confirms that stochastic modeling remains critical in real-world, high-dimensional settings.
Moreover, models operating in VAE latent spaces perform worse than ours using strong semantic latents. We attribute this to the limited semantic fidelity of low-dimensional VAE representations, which are aligned worse with downstream perception tasks such as detection.
To disentangle whether the VAE baseline underperforms due to weaker future modeling or a less semantic feature space, we re-encode VAE predictions into DINOv3 space before detection (VAE decode $\to$ DINOv3 encode). Re-encoding does not help and slightly hurts performance (17.5 $\to$ 16.5 AP$_{\text{L}}$(3)), because the pixel roundtrip introduces reconstruction artifacts that degrade the DINOv3 features. This confirms that the information is already lost at the prediction stage, and no post-hoc re-encoding can recover it.
\ourmodel{} also outperforms the prior stochastic world model~\cite{walker2025generalist}, validating the effectiveness of the proposed design.

\noindent\textbf{Depth estimation.}
In the right block of Table~\ref{tab:future_detections_joint}, we further extend our evaluation to depth estimation, another downstream perception task, to assess the generality of the predicted latents. Again, \ourmodel{} outperforms all baselines across all depth metrics, confirming that the predicted latents remain effective across diverse downstream tasks.

\noindent\textbf{Pixel generation.}
We additionally evaluate the quality of decoded future frames using Fréchet Video Distance (FVD)~\citep{unterthiner2018towards} (the lower the better), although pixel generation is not the primary focus of this work. 
To this end, we train a lightweight pixel decoder from DINOv3 latents and compute FVD on the decoded future frames. \ourmodel{} achieves an FVD of $87.3$, corresponding to a $43\%$ improvement over the deterministic predictor ($152.4$), confirming that our predicted futures are distributionally closer to real driving videos.

\begin{figure}[t]
  \vskip 0.2in
  \begin{center}
    \includegraphics[width=0.95\textwidth,height=0.23\textheight,keepaspectratio=false]{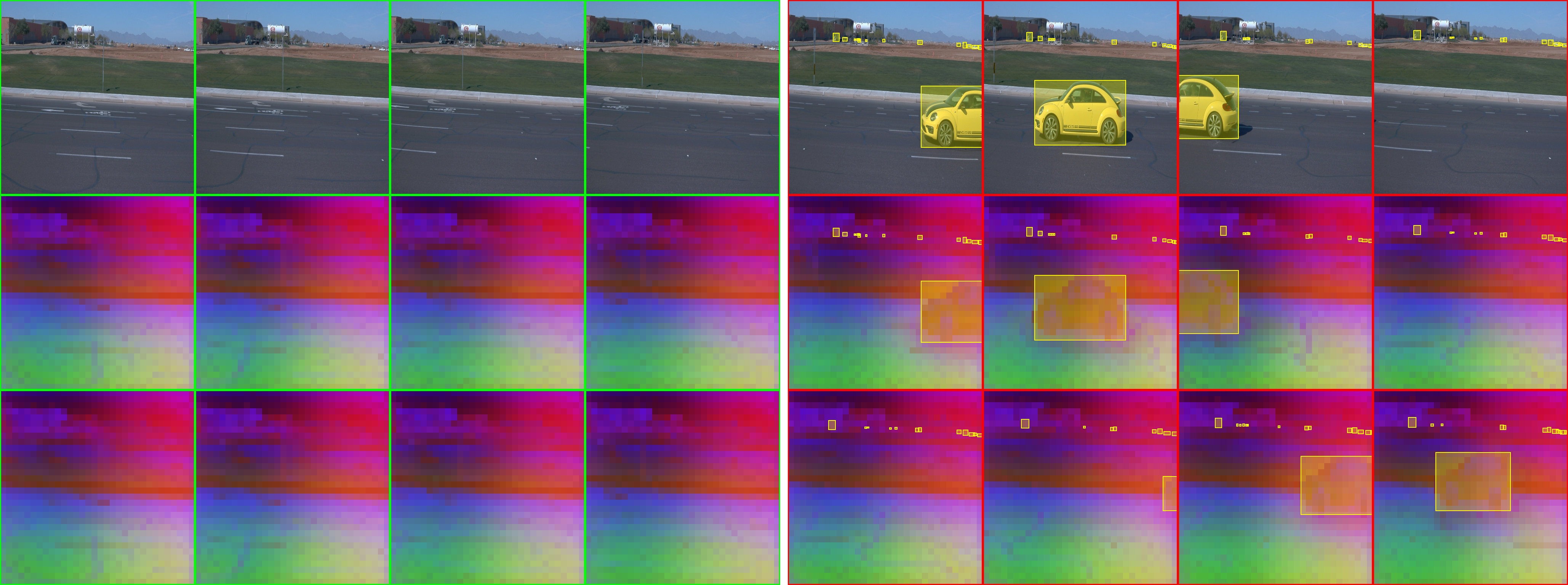}
    \caption{
    \textbf{Qualitative examples on our \ourtask{} benchmark.}
    Top row: Ground truth frames and oracle detections.
    Middle and bottom rows: PCA visualizations of predicted latents with overlaid detections for two distinct rollouts.
    The middle row shows the model anticipating an oncoming vehicle. The bottom row depicts a prediction where a different vehicle appears later.
    Green borders indicate context frames; red borders indicate predicted future frames.}
    \label{fig:waymo_visual_samples_multi_predictions}
  \end{center}
  \vspace{-0.5cm}
\end{figure}

\noindent\textbf{Oracle comparison.}
For reference, we also include an Oracle that applies the downstream predictors to ground-truth future latents. The remaining gap between \ourmodel{} and the Oracle highlights the inherent difficulty of real-world future prediction. Nevertheless, qualitative results in Figure~\ref{fig:waymo_visual_samples_multi_predictions} show that \ourmodel{} produces diverse and plausible futures.


\begin{table}[t]
\centering
\caption{
\textbf{Object detection and depth estimation performance on \ourtask{}.}
}
\label{tab:future_detections_joint}
\resizebox{\linewidth}{!}{
\begin{tabular}{l c cc cccc}
\toprule
\multirow{2}{*}{Method} & \multirow{2}{*}{Backbone} & \multicolumn{2}{c}{Object Detection} & \multicolumn{4}{c}{Depth Prediction} \\
\cmidrule(lr){3-4} \cmidrule(lr){5-8}
& & $\mathrm{AP}_{\mathrm{L}}(3)\uparrow$ & $\mathrm{AP}_{\mathrm{L}}(6)\uparrow$ & RMSE$\downarrow$ & $\delta_1\uparrow$ & $\delta_2\uparrow$ & $\delta_3\uparrow$ \\
\midrule
DINO-WM & DINOv3 & 14.5 & 14.5 & 0.12 & 0.51 & 0.76 & 0.88 \\
Deterministic Predictor & DINOv3 & 15.2 & 15.2 & 0.11 & 0.54 & 0.78 & 0.89 \\
WAN 2.2 VAE WM & Wan 2.2 VAE & 17.5 & 18.2 & 0.10 & 0.57 & 0.80 & 0.90 \\
WAN 2.2 VAE WM (re-encoded) & DINOv3 & 16.5 & 17.0 & 0.107 & 0.54 & 0.78 & 0.89 \\
\cite{walker2025generalist} & DINOv3 & 16.5 & 17.1 & 0.102 & 0.56 & 0.79 & 0.90 \\
\textbf{FlowWM (Ours)} & DINOv3 & \textbf{20.9} & \textbf{21.7} & \textbf{0.078} & \textbf{0.723} & \textbf{0.858} & \textbf{0.913} \\
\midrule
Oracle & DINOv3 & \textit{65.1} & \textit{65.1} & \textit{0.043} & \textit{0.854} & \textit{0.951} & \textit{0.977} \\
\bottomrule
\end{tabular}
}
\vspace{-1em}
\end{table}

\subsection{Ablations}
\label{sec:ablation}

\begin{wrapfigure}{r}{0.5\linewidth}
\vspace{-1.5em}
\centering
\includegraphics[width=1.\linewidth]{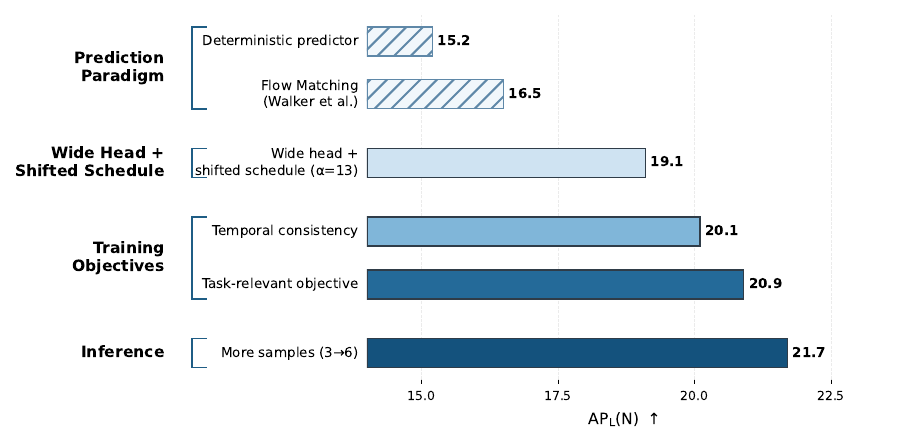}
\caption{\textbf{Roadmap for \ourmodel}, measured by downstream object detection performance on \ourtask{} benchmark.}
\label{fig:ablation_roadmap}
\vspace{-1.5em}
\end{wrapfigure}

We perform extensive ablations on~\ourtask{} to verify the key design choices of our~\ourmodel{} model.
Figure~\ref{fig:ablation_roadmap} shows the overall roadmap of key ingredients, while more detailed results are presented in Appendix~\ref{sec:appendix_ablations}.

\noindent\textbf{Depth and width scaling.}
We study the effect of model depth and width in latent world models.
Varying model depth from 2 to 8 DiT layers (Figure~\ref{fig:depth_width_ablations}a) yields no  improvement in downstream detection performance, 
indicating that model depth is not a limiting factor for the task.
In contrast, varying the width of the projection head (Figure~\ref{fig:depth_width_ablations}b) has a pronounced effect: narrow heads severely underperform, while increasing width yields improvements up to our chosen depth $d=1024$, beyond which gains saturate.
This suggests that the width should be at least as large as the latent dimension, aligning with findings in the image domain~\citep{zheng2025diffusiontransformersrepresentationautoencoders}.


\noindent\textbf{Temporal consistency.}
In Table~\ref{tab:ablation_loss}, we ablate the temporal consistency loss (Eq.~(\ref{eqn:temporal_consistency_loss})).
It improves by 5.2\% relatively to the baseline, indicating that enforcing coherent temporal dynamics benefits downstream perception.
Moreover, incorporating this loss via the proposed one-step projection incurs no additional computational overhead during training.

\noindent\textbf{Task-driven objective.}
We also ablate the downstream task-driven objective, in which the detector loss is backpropagated during training on the \ourtask{} benchmark.
This provides additional performance gains.
We hypothesize that these gains are bounded by the limited robustness of the frozen detector to noisy predicted latents, which results in weaker gradients.
Moreover, because the quality of the projected endpoint latent varies across flow-matching timesteps, appropriately weighting the task-driven loss over timesteps is important (Table~\ref{tab:detector_schedule}).

\noindent\textbf{Shifting of timestep schedule.}
In Table~\ref{tab:waymo_time_shift}, we show that skewing the training timestep distribution (Section~\ref{sec: timestep_shifting}) toward noisier timesteps is crucial.
While this strategy was originally introduced to facilitate high-resolution image generation~\citep{peebles2023dit}, our results reveal that the same principle applies when increasing latent dimension and prediction horizon in video world modeling.

\noindent\textbf{Scaling with number of samples.}
\label{sec:sampling_budget}
One key advantage of stochastic world models is the ability to sample multiple futures.
We study how detection performance scales with the number of sampled futures $N$.
Figure~\ref{fig:sampling-budget} shows \textbf{AP$_{\text{L}}$(N)} as a function of the sampling budget.
The performance improves consistently as $N$ increases. 
In contrast, deterministic world models produce the same future on the same context, and thus provide no benefit from additional samples.

\begin{figure}[t]
  \begin{center}
    \includegraphics[width=0.5\linewidth]{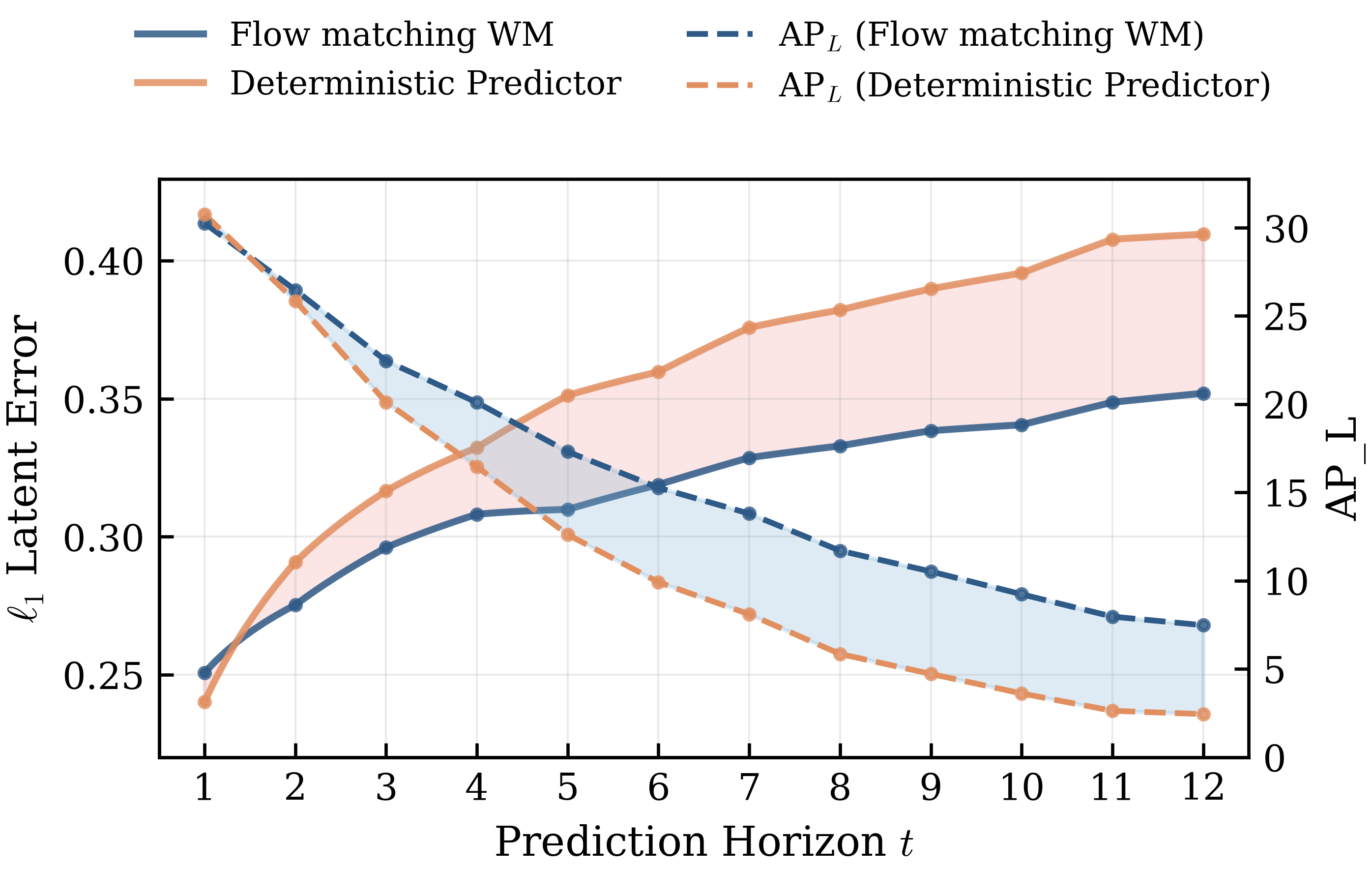}
    \caption{
    \textbf{$\ell_1$ latent error and $\mathrm{AP}_L$ as a function of prediction horizon.}
      Deterministic predictors exhibit faster error growth and $\mathrm{AP}_L$ degradation, while our stochastic model maintains lower error and higher performance over long horizons.
    }
    \label{fig:temporal-error}
  \end{center}
\end{figure}

\subsection{Temporal Error Analysis}
\label{sec:temporal_error}

As deterministic predictors learn to predict the mean of possible futures, while short-term predictions can be accurate, they tend to degrade quickly as uncertainty compounds over time.
In contrast, stochastic world models can represent multiple plausible futures and are therefore expected to degrade more slowly as the forecasting horizon grows.

To test this, we analyze how prediction performance evolves as a function of the number of predicted frames.
Figure~\ref{fig:temporal-error} reports the $\ell_1$ latent error across the prediction horizon. For the stochastic model, we sample 3 trajectories and report the performance of the closest to the ground-truth, similar to our detection metric. As expected,
both models exhibit increasing error over time, but the deterministic predictor degrades much faster, as it strays further from a valid mode, while the stochastic one keeps predicting a valid future. 
The figure also shows that the latent reconstruction error directly correlates with the evolution of the detection performance \text{AP$_{\text{L}}$(N)}, which exhibits a similar widening gap over time.



These results demonstrate that deterministic world models accumulate error more rapidly in latent space, which in turn degrades downstream task performance. Stochastic world models mitigate this failure mode by preserving diverse future hypotheses, resulting in more stable and useful predictions over
long horizons.

\section{Conclusion}

This work shows that accurate future prediction in latent space requires both stochastic modeling and semantically rich representations. Deterministic predictors collapse multimodal futures and degrade quickly over long horizons, and low-dimensional VAE latents lack the semantic structure needed for downstream perception. 
To address this, we propose~\ourmodel{}, a stochastic Flow Matching model which operates directly in pretrained visual features and benefits from temporal and task-driven objectives.
We further introduce a synthetic benchmark and a real-world \ourtask{} benchmark. 
Our model achieves better mode coverage, slower horizon degradation, and stronger downstream detection performance, highlighting the importance of the proposed design choices.
Discussions on limitation and broader impact are presented in the Appendix~\ref{sec:appendix_limitation} and~\ref{sec:broader_impact}.



\section*{Acknowledgements}
S. Chen was funded in part by the French government under management of Agence Nationale de la Recherche as part of the ``France 2030'' program, reference ANR-23-IACL-0008 (PR[AI]RIE-PSAI projet) and the ANR project 3D-GEM (ANR-25-CE23-7777-01). K. Alahari was supported in part by the Institute of Information \& Communications Technology Planning \& Evaluation (IITP) grant funded by the Korean Government (MSIT) (No. RS-2024-00457882, National AI Research Lab Project).

\clearpage
\bibliographystyle{assets/plainnat}
\bibliography{references}

\newpage
\beginappendix


\section{Multimodal Futures Require Stochastic Models}
\label{sec:appendix_stochastic_necessity}

\begin{wrapfigure}{r}{0.4\linewidth}
\vspace{-1em}
\centering
\includegraphics[width=1\linewidth]{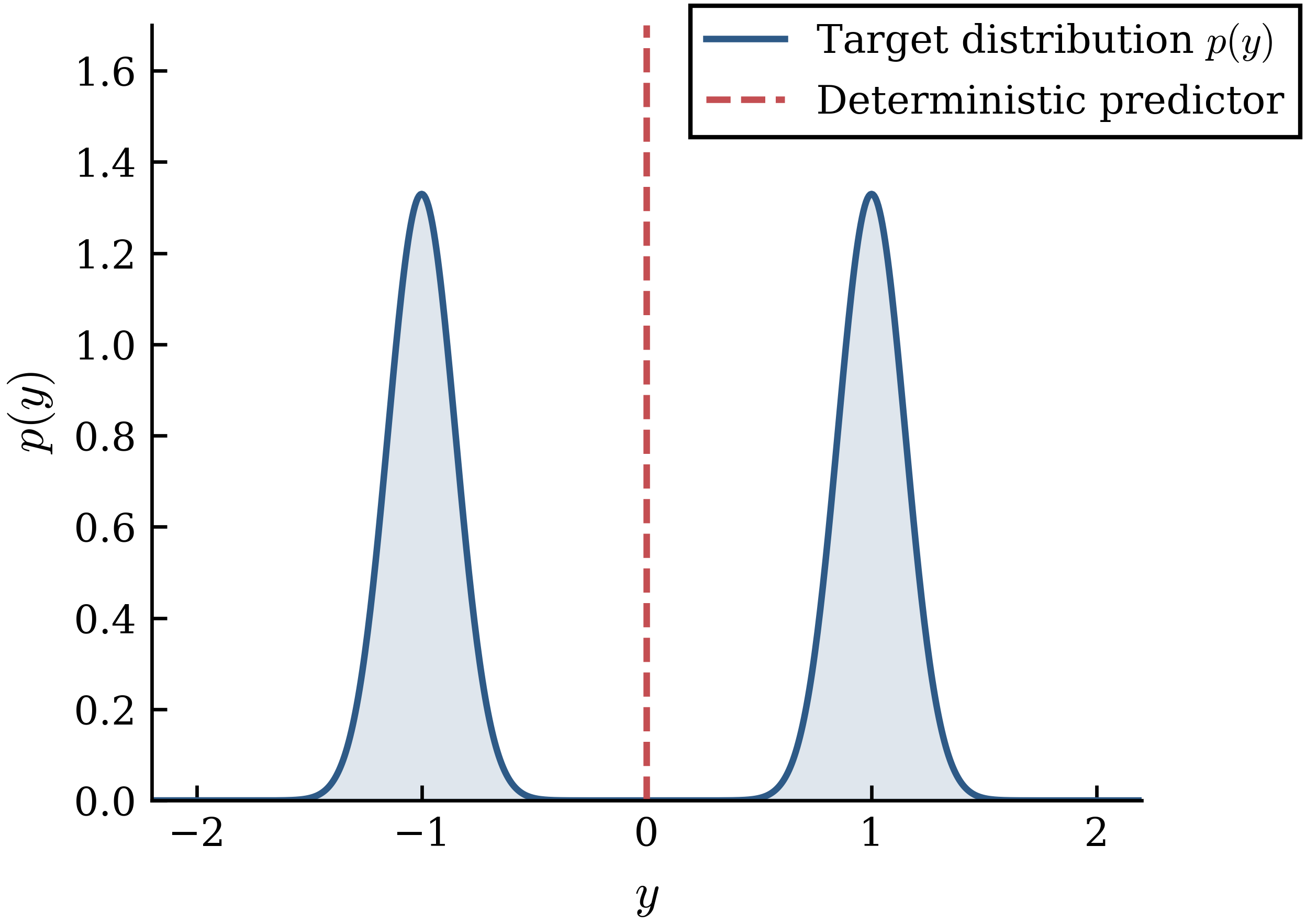}
\caption{In a bimodal target distribution, deterministic predictors trained with $\ell_1$ or $\ell_2$ losses collapse to a single point estimate (the conditional mean or median), which lies between modes and corresponds to no valid future. }
\label{fig:bimodal_distribution}
\vspace{-2em}
\end{wrapfigure}

Future states are often intrinsically multimodal: agents may take different actions, and objects may move differently due to environmental uncertainty.
As a result, a core requirement of world models is to capture multiple plausible futures conditioned on the same past.
However, most existing latent-space based world models~\citep{zhou2024dinowm,karypidis2024dinof} rely on deterministic predictors trained with pointwise regression losses, which fundamentally cannot represent multimodal dynamics.

Consider a simple bimodal future: conditioned on the same context, the system evolves according to a fair coin toss:
\begin{equation}
    Y =
\begin{cases}
y_L, & \text{with probability } \tfrac12,\\[1mm]
y_R, & \text{with probability } \tfrac12,
\end{cases}
\qquad y_L \neq y_R,
\end{equation}

A deterministic predictor trained with an $\ell_2$ loss outputs the conditional mean
$\mathbb{E}[Y \mid X_t = x_t] = \tfrac12 (y_L + y_R)$, which lies between modes and corresponds to no valid future (Figure~\ref{fig:bimodal_distribution}).

More generally, deterministic world models trained with $\ell_1$ or $\ell_2$ losses converge to the conditional mean or median, averaging over all plausible futures rather than representing any valid mode of the distribution.
%


\paragraph{Formal analysis.} Deterministic predictors trained with an $\ell_1$ or $\ell_2$ loss solve a
\emph{conditional regression} problem.
Given a context $X_t = x_t$, the model outputs a single value $\hat{y}(x_t)$ and minimizes a risk functional.
We analyze both cases below for the bimodal distribution where $Y = y_L$ or $Y = y_R$ each with probability $\tfrac12$.

\noindent\textbf{Squared loss ($\ell_2$).}
The $\ell_2$ risk is
\[
R_2(\hat{y})
=
\tfrac12 (y_L-\hat{y})^2 + \tfrac12 (y_R-\hat{y})^2.
\]
Setting the derivative to zero yields
\[
\hat{y}^\star = \tfrac12(y_L + y_R).
\]
More generally, the minimizer of the squared loss $\mathcal{R}(\hat{y}) = \mathbb{E}[\|\hat{y}(X_t) - Y\|_2^2]$ is the conditional expectation:
\[
\hat{y}^\star(x_t)
=
\mathbb{E}[\,Y \mid X_t = x_t\,].
\]
Thus, any deterministic predictor returns a \emph{single} point estimate equal to the conditional mean over all plausible futures.

\noindent\textbf{Absolute loss ($\ell_1$).}
Similarly, the $\ell_1$ risk is
\[
R_1(\hat{y})
=
\tfrac12 |y_L - \hat{y}| + \tfrac12 |y_R - \hat{y}|,
\]
whose minimizers are the medians of $Y$; in particular, the midpoint
$\tfrac12(y_L + y_R)$ is always optimal.

In both cases, the predictor returns a single point estimate that averages over all plausible futures, yielding an output that does not correspond to any valid mode of the distribution.

This failure mode becomes worse in \emph{high-dimensional latent space world models}, where averaging destroys semantic structure and temporal compounding amplifies errors.
Thus, stochastic generative modeling is essential for robust world modeling.

\section{Bouncing Shapes Benchmark}
\label{sec:appendix_bouncing_shapes_benchmark}

\subsection{Bouncing Shapes Modularity}

\begin{figure}[t]
  \centering
  \includegraphics[width=0.5\linewidth]{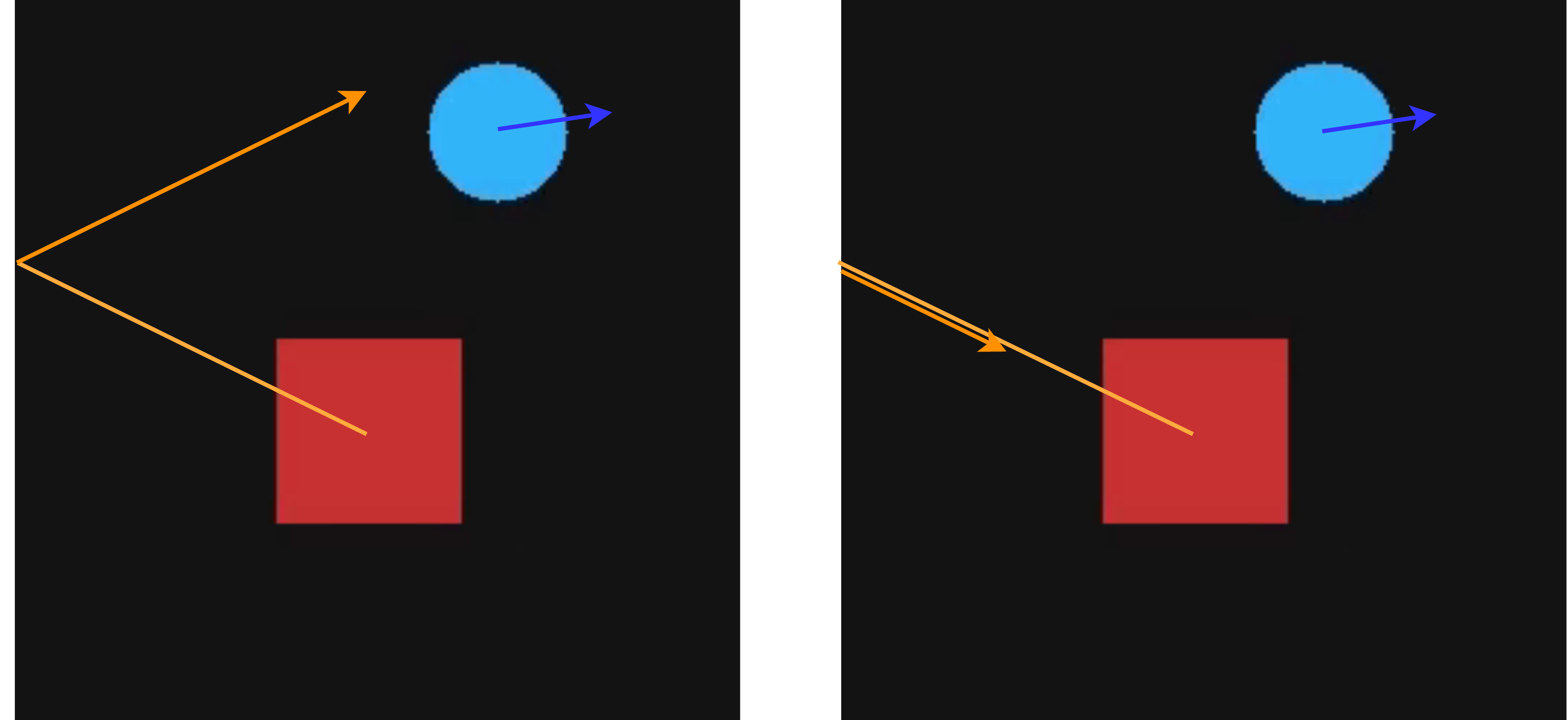}
  \caption{\textbf{Bouncing Shapes dataset.} Two objects move in a 2D box, with stochastic bounces at wall collisions. At each collision, the object either performs a standard bounce or reverses its velocity with probability $0.5$, creating exponentially branching futures.}
  \label{fig:bouncing_shapes}
\end{figure}

\begin{figure}[t]
  \centering
  \begin{subfigure}{1\linewidth}
        \centering
        \includegraphics[width=\linewidth]{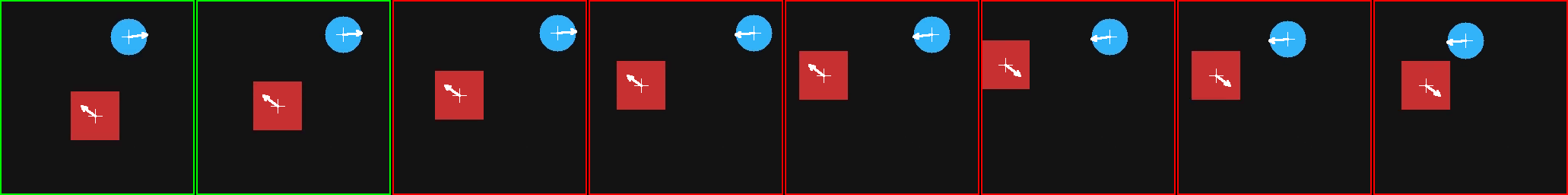}
        \caption{Trajectory 1.}
        \label{fig:bouncing_shapes_traj_1}
    \end{subfigure}

    \begin{subfigure}{1\linewidth}
        \centering
        \includegraphics[width=\linewidth]{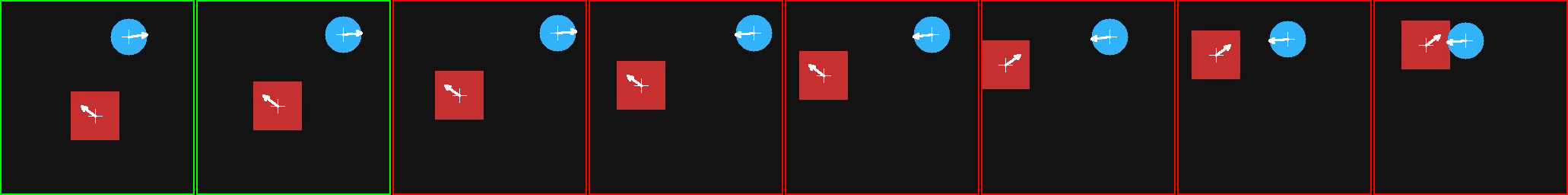}
        \caption{Trajectory 2.}
        \label{fig:bouncing_shapes_traj_2}
    \end{subfigure}
  \caption{\textbf{Example trajectories in the Bouncing Shapes benchmark.} Green box denotes the context frame, and red box denotes the target frame for prediction.}
  \label{fig:bouncing_shapes_trajs}
\end{figure}

The dataset is fully editable and can be made arbitrarily complex. It is possible to change the number, color, shapes of objects, and use arbitrarily complex distributions for rebounds and trajectories.
Figure~\ref{fig:bouncing_shapes} and ~\ref{fig:bouncing_shapes_trajs} provide examples in the Bouncing Shapes benchmark.

\subsection{Metrics}
\label{sec:bouncing_shapes_metrics}

\begin{algorithm}[H]
\caption{Precision Error}
\label{alg:precision_error}
\begin{algorithmic}
  \STATE {\bfseries Input:} predictions $\mathcal{P} = \{p_i\}_{i=1}^N$, ground-truth futures $\mathcal{G} = \{g_j\}_{j=1}^M$, distance $\ell(\cdot, \cdot)$
  \STATE {\bfseries Output:} precision error
  \STATE Initialize $\mathcal{E} \leftarrow \emptyset$. \hfill{\itshape store matched prediction errors}
  \FOR{each $p \in \mathcal{P}$}
    \STATE $e_p \leftarrow \min\limits_{g \in \mathcal{G}} \ell(p, g)$. \hfill{\itshape match closest ground truth}
    \STATE $\mathcal{E} \leftarrow \mathcal{E} \cup \{e_p\}$. \hfill{\itshape add error to the set}
  \ENDFOR
  \STATE \textbf{return} $\frac{1}{|\mathcal{E}|} \sum_{e \in \mathcal{E}} e$. \hfill{\itshape average over errors}
\end{algorithmic}
\end{algorithm}

\begin{algorithm}[H]
\caption{Recall Error}
\label{alg:recall_error}
\begin{algorithmic}
  \STATE {\bfseries Input:} predictions $\mathcal{P} = \{p_i\}_{i=1}^N$, ground-truth futures $\mathcal{G} = \{g_j\}_{j=1}^M$, distance $\ell(\cdot, \cdot)$
  \STATE {\bfseries Output:} recall error
  \STATE Initialize $\mathcal{E} \leftarrow \emptyset$. \hfill{\itshape store matched prediction errors}
  \FOR{each $g \in \mathcal{G}$}
    \STATE $e_g \leftarrow \min\limits_{p \in \mathcal{P}} \ell(p, g)$. \hfill{\itshape match closest prediction}
    \STATE $\mathcal{E} \leftarrow \mathcal{E} \cup \{e_g\}$. \hfill{\itshape add error to the set}
  \ENDFOR
  \STATE \textbf{return} $\frac{1}{|\mathcal{E}|} \sum_{e \in \mathcal{E}} e$. \hfill{\itshape average over errors}
\end{algorithmic}
\end{algorithm}

\paragraph{F1 Error.}
We define the F1 error as the harmonic mean of Precision Error ($E_{\text{prec}}$) and Recall Error ($E_{\text{rec}}$):

\begin{equation}
E_{\text{F1}}
\;=\;
2 \cdot \frac{E_{\text{prec}} \cdot E_{\text{rec}}}{E_{\text{prec}} + E_{\text{rec}}}.
\end{equation}

\section{\ourtask{} Benchmark}
\label{sec:appendix_waymo_benchmark}

\begin{figure}[h]
  \vskip 0.2in
  \begin{center}
    \includegraphics[width=0.95\textwidth]{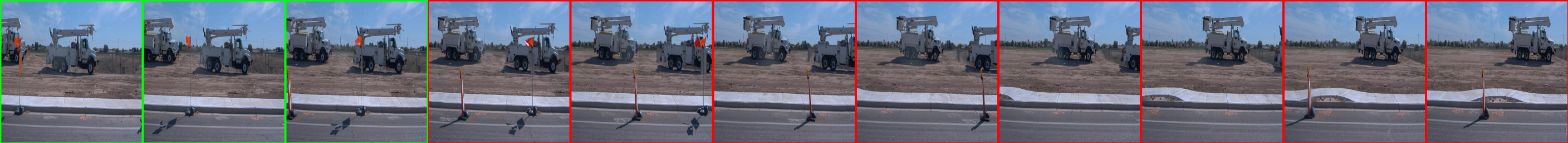}
    \vspace{0.25em}
    \includegraphics[width=0.95\textwidth]{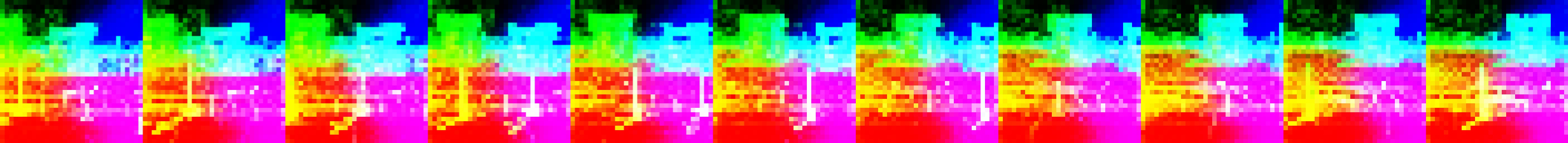}
    \vspace{0.25em}
    \includegraphics[width=0.95\textwidth]{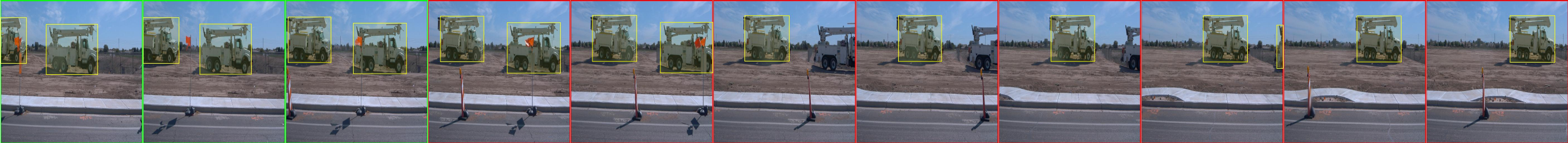}
    \caption{Waymo visual samples: original video, ground-truth latents, and oracle predictions. Green border indicates context; red border indicates future prediction.}
    \label{fig:waymo_visual_samples_appendix_4}
  \end{center}
\end{figure}

\begin{figure}[h]
  \vskip 0.2in
  \begin{center}
    \includegraphics[width=0.95\textwidth]{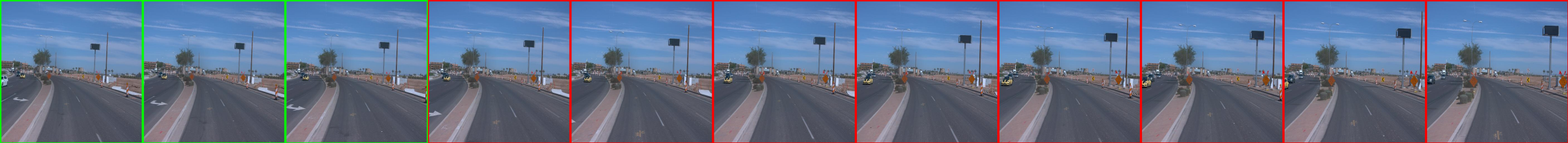}
    \vspace{0.25em}
    \includegraphics[width=0.95\textwidth]{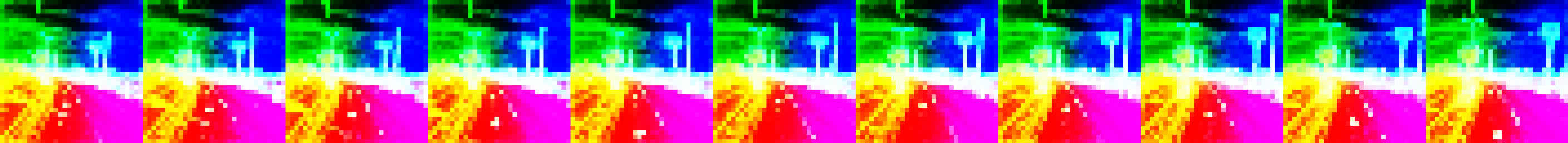}
    \vspace{0.25em}
    \includegraphics[width=0.95\textwidth]{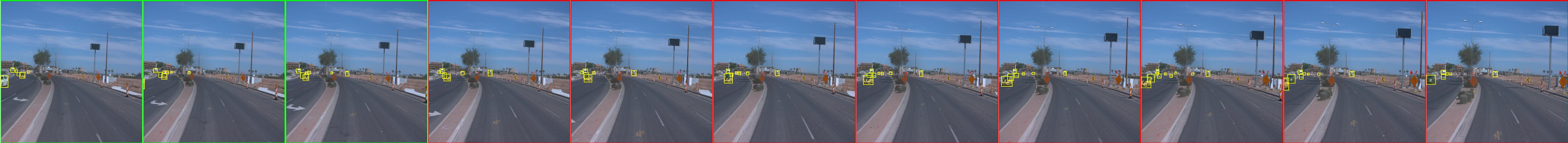}
    \caption{Waymo visual samples: original video, ground-truth latents, and oracle predictions. Green border indicates context; red border indicates future prediction.}
    \label{fig:waymo_visual_samples_appendix_6}
  \end{center}
\end{figure}

\begin{figure}[h]
  \vskip 0.2in
  \begin{center}
    \includegraphics[width=0.95\textwidth]{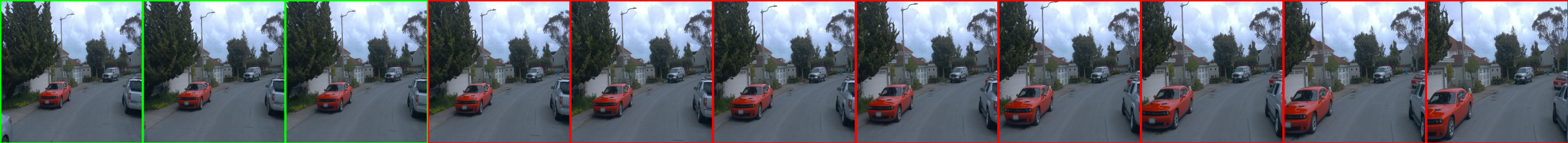}
    \vspace{0.25em}
    \includegraphics[width=0.95\textwidth]{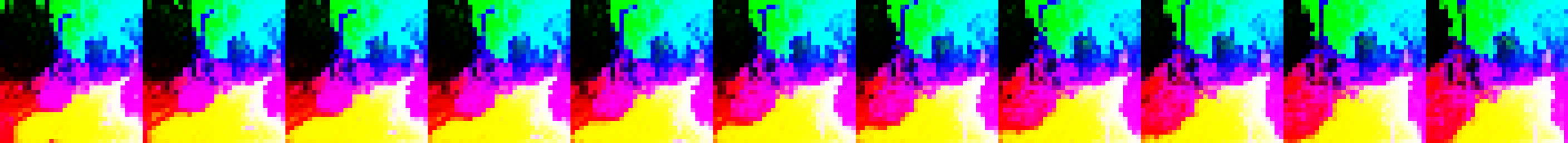}
    \vspace{0.25em}
    \includegraphics[width=0.95\textwidth]{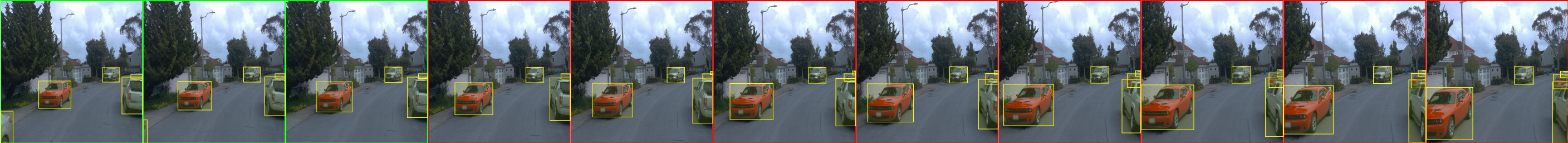}
    \caption{Waymo visual samples: original video, ground-truth latents, and oracle predictions. Green border indicates context; red border indicates future prediction.}
    \label{fig:waymo_visual_samples_appendix_10}
  \end{center}
\end{figure}

\begin{figure}[h]
  \vskip 0.2in
  \begin{center}
    \includegraphics[width=0.95\textwidth]{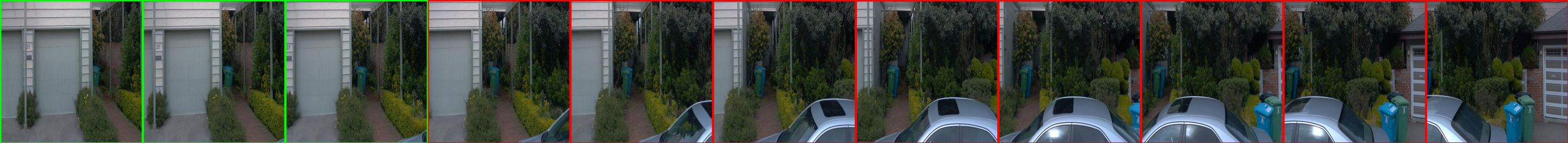}
    \vspace{0.25em}
    \includegraphics[width=0.95\textwidth]{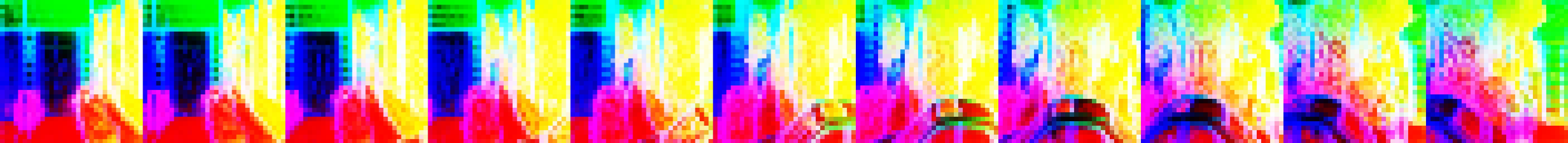}
    \vspace{0.25em}
    \includegraphics[width=0.95\textwidth]{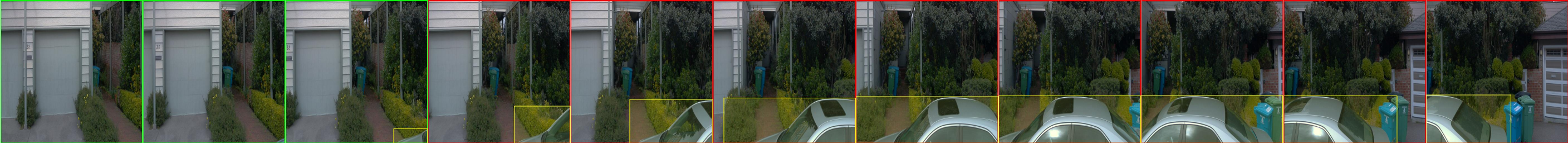}
    \caption{Waymo visual samples: original video, ground-truth latents, and oracle predictions. Green border indicates context; red border indicates future prediction.}
    \label{fig:waymo_visual_samples_appendix_12}
  \end{center}
\end{figure}

\begin{figure}[h]
  \vskip 0.2in
  \begin{center}
    \includegraphics[width=0.95\textwidth]{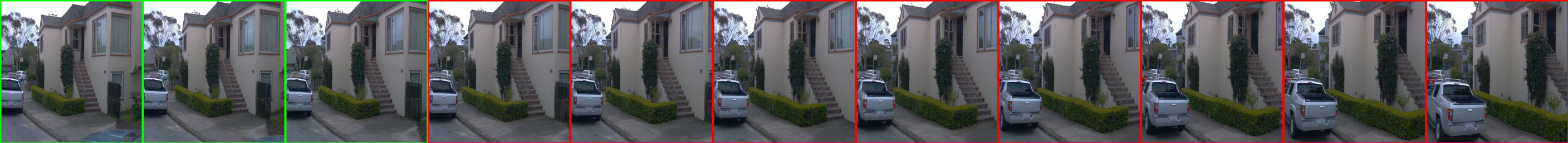}
    \vspace{0.25em}
    \includegraphics[width=0.95\textwidth]{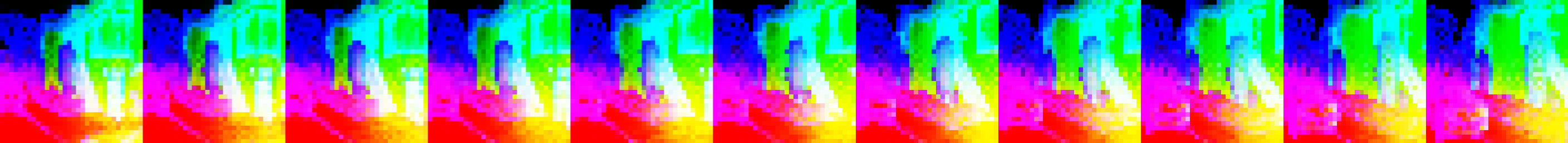}
    \vspace{0.25em}
    \includegraphics[width=0.95\textwidth]{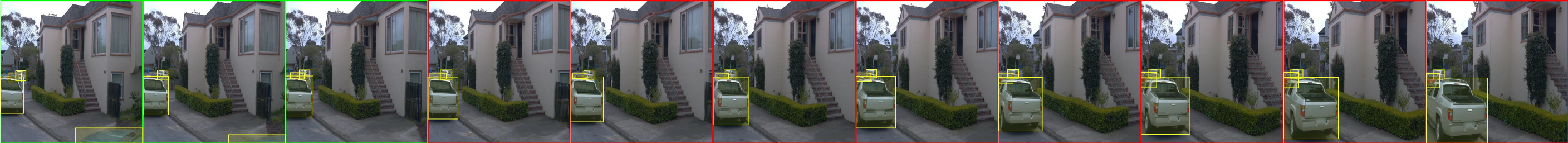}
    \caption{Waymo visual samples: original video, ground-truth latents, and oracle predictions. Green border indicates context; red border indicates future prediction.}
    \label{fig:waymo_visual_samples_appendix_13}
  \end{center}
\end{figure}

Figure~\ref{fig:waymo_visual_samples_appendix_4} to ~\ref{fig:waymo_visual_samples_appendix_13} show visual examples in our \ourtask{} Benchmark for the downstream object detection task.

\section{Detailed Ablation Results}
\label{sec:appendix_ablations}

We perform extensive ablations on~\ourtask{} object detection downstream task to verify the key design choices of our~\ourmodel{} model.

\begin{figure}[h]
    \centering
    \includegraphics[width=1\linewidth]{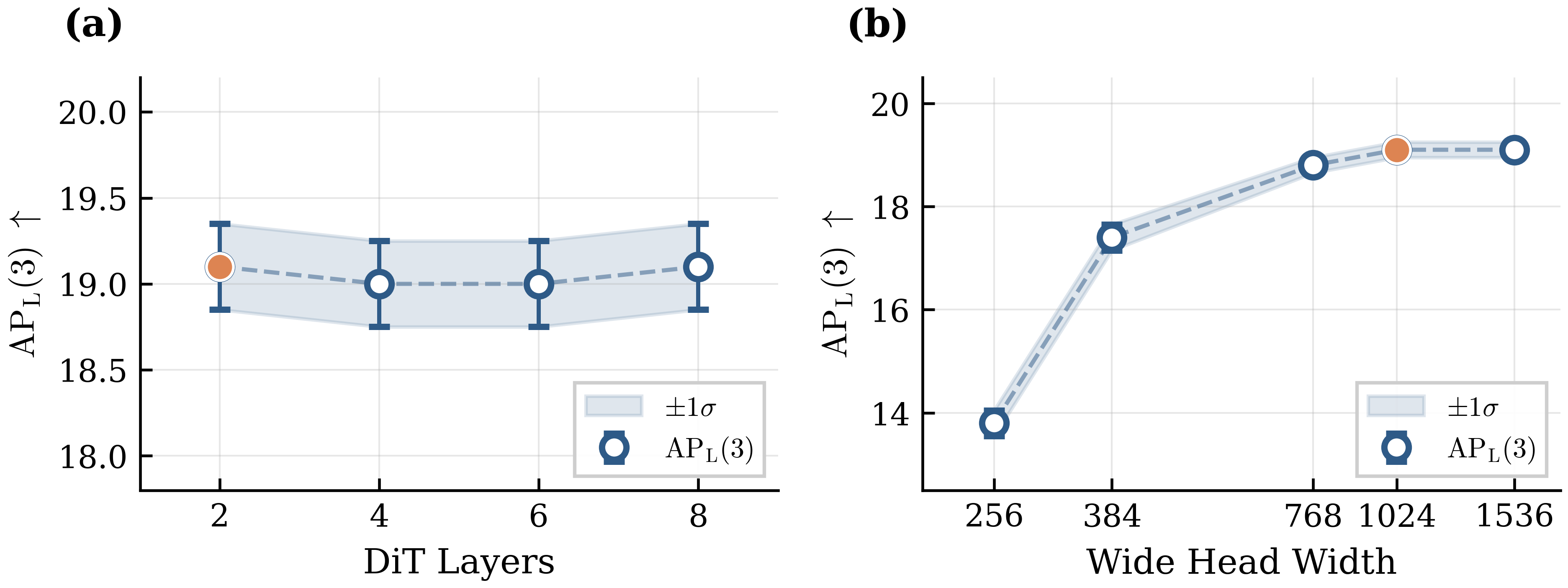}
    \caption{Ablation of model sizes on Waymo Open Dataset: depth of DiT layers (a) and width of the projection head (b). 
    Orange indicates default configuration. 
    Mean $\pm$ std over 3 runs.
    }
    \label{fig:depth_width_ablations}
\end{figure}

\noindent\textbf{Depth and width scaling.}
We study the effect of model depth and width in latent world models.
Varying model depth from 2 to 8 DiT layers (Figure~\ref{fig:depth_width_ablations}a) yields no improvement in downstream detection performance, 
indicating that model depth is not a limiting factor for the task.
In contrast, varying the width of the projection head (Figure~\ref{fig:depth_width_ablations}b) has a pronounced effect: narrow heads severely underperform, while increasing width yields improvements up to our chosen depth $d=1024$, beyond which gains saturate.
This suggests that the width should be at least as large as the latent dimension, aligning with findings in the image domain~\citep{zheng2025diffusiontransformersrepresentationautoencoders}.

\noindent\textbf{Temporal consistency.}
In Table~\ref{tab:ablation_loss}, we ablate the temporal consistency loss (Eq.~(\ref{eqn:temporal_consistency_loss})).
It improves by 5.2\% relatively to the baseline, indicating that enforcing coherent temporal dynamics benefits downstream perception.
Moreover, incorporating this loss via the proposed one-step projection incurs no additional computational overhead during training.

\begin{table}[h]
\centering
  \caption{
    \textbf{Ablation of training objectives on~\ourtask{} benchmark.}
  }
  \label{tab:ablation_loss}
  \begin{center}
    \setlength{\tabcolsep}{6pt}
    \begin{small}
      \renewcommand{\arraystretch}{1.2}
      \begin{tabular}{@{}lc@{}}
        \toprule
        Method & \textbf{AP$_{\text{L}}$(3)} $\uparrow$ \\
        \midrule
        Baseline & 19.1 \\
        + Temporal consistency loss & 20.1 \\
        + Detector loss & \textbf{20.9} \\
        \bottomrule
      \end{tabular}
    \end{small}
  \end{center}
\end{table}

\noindent\textbf{Task-driven objective.}
We also ablate the downstream task-driven objective, in which the detector loss is backpropagated during training on the \ourtask{} benchmark.
This provides additional performance gains, though smaller than those from the more general temporal consistency loss.
We hypothesize that these gains are bounded by the limited robustness of the frozen detector to noisy predicted latents, which results in weaker gradients.
Moreover, because the quality of the projected endpoint latent varies across flow-matching timesteps, appropriately weighting the task-driven loss over timesteps is important.

\begin{wraptable}{r}{0.2\linewidth}
\centering
  \caption{\ourtask{}: detector loss schedule ablation.}
  \label{tab:detector_schedule}
  \begin{center}
    \setlength{\tabcolsep}{4pt}
    \begin{small}
      \renewcommand{\arraystretch}{1.2}
      \begin{tabular}{@{}ll@{}}
        \toprule
        $\gamma$ & \textbf{AP$_{\text{Large}}$(3)} $\uparrow$ \\
        \midrule
        1 & 12.4 \\
        7 & 19.7 \\
        12 & 20.5 \\
        15 & 20.9 \\
        20 & 20.7 \\
        \bottomrule
      \end{tabular}
    \end{small}
  \end{center}

\end{wraptable}

As shown in Table~\ref{tab:detector_schedule}, increasing $\gamma$ substantially improves detection performance, indicating that backpropagating the detector loss uniformly across timesteps overconstrains the Flow Matching objective. Emphasizing supervision at later timesteps allows the model to preserve a well-structured global flow while injecting task-specific constraints near the endpoint. However, excessively large values of $\gamma$ lead to saturation and slight degradation in performance, suggesting a trade-off between task specialization and generative flexibility.

\noindent\textbf{Shifting of timestep schedule.}
In Table~\ref{tab:waymo_time_shift}, we show that skewing the training timestep distribution (Section~\ref{sec: timestep_shifting}) toward noisier timesteps is crucial.
While this strategy was originally introduced to facilitate high-resolution image generation~\citep{peebles2023dit}, our results reveal that the same principle applies when increasing latent dimension and prediction horizon in video world modeling.

\begin{table}[h]
    \caption{\textbf{Effect of timestep schedule shifting.} $\alpha$ is the timestep skew parameter. Biasing the training distribution toward noisier timesteps substantially improves performance.}
    \label{tab:waymo_time_shift}
    \begin{center}
        \setlength{\tabcolsep}{4pt}
        \begin{small}
            \renewcommand{\arraystretch}{1.2}
            \begin{tabular}{@{}lcccccc@{}}
                \toprule
                $\alpha$ & 1 & 7 & 11 & \textbf{13} & 15 & 17 \\
                \midrule
                \textbf{AP$_{\text{L}}$(3)} $\uparrow$ & 14.3 & 18.1 & 18.7 & \textbf{19.1} & 19.0 & 18.7 \\
                \bottomrule
            \end{tabular}
        \end{small}
    \end{center}
    \vskip -0.1in
\end{table}

\begin{figure}[h]
  \centering
  \includegraphics[width=0.5\linewidth]{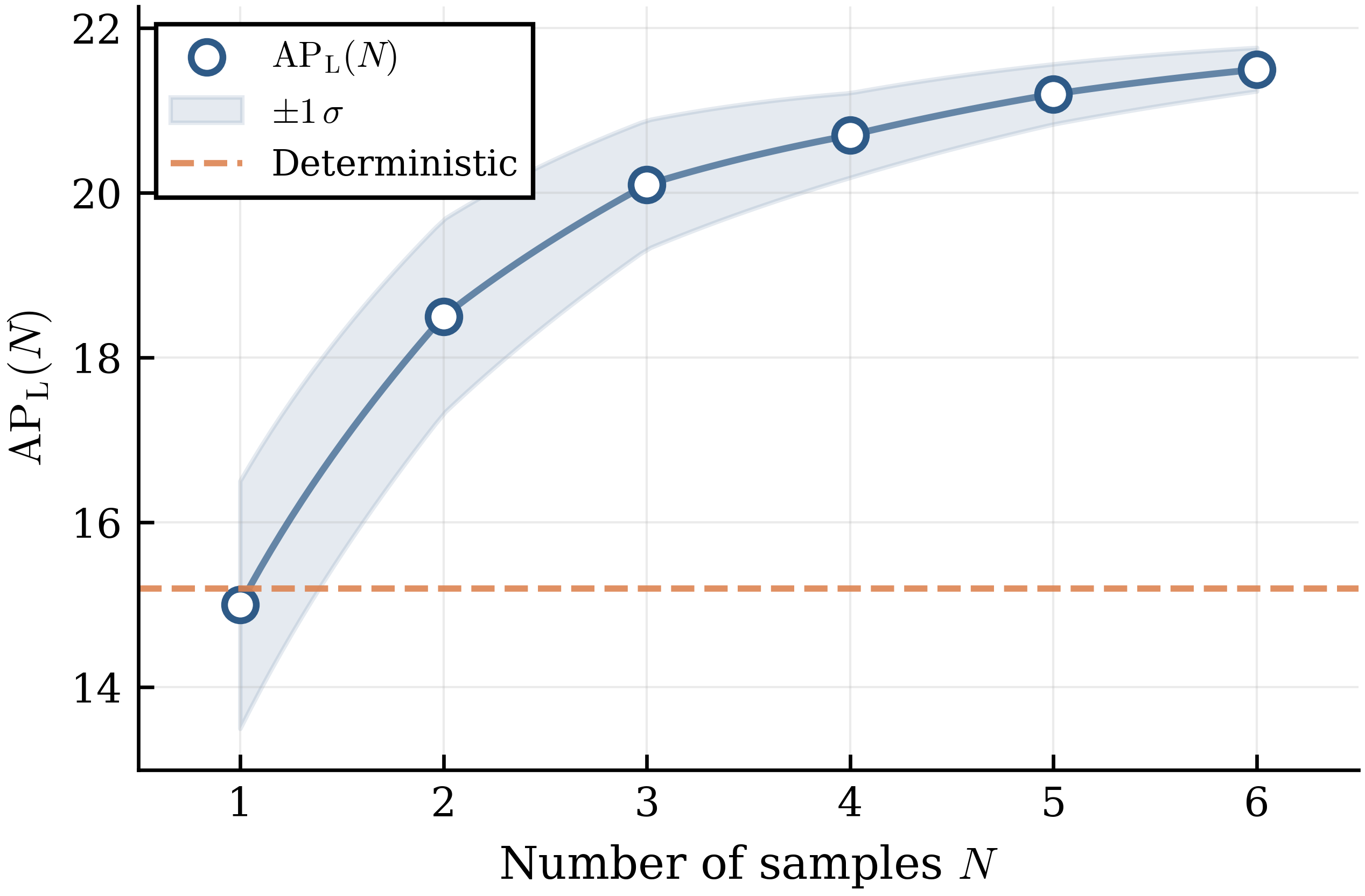}
  \caption{
    \textbf{Sampling budget.} AP$_{\text{L}}$(N) as a function of the number
    of sampled futures. Performance improves monotonically with sampling budget
    for the stochastic world model, while deterministic predictors show no gain.}
  \label{fig:sampling-budget}
\end{figure}

\noindent\textbf{Scaling with number of samples.}
One key advantage of stochastic world models is the ability to sample multiple futures.
We study how detection performance scales with the number of sampled futures $N$.
Figure~\ref{fig:sampling-budget} shows \textbf{AP$_{\text{L}}$(N)} as a function of the sampling budget.
The performance improves consistently as $N$ increases. 
In contrast, deterministic world models produce the same future on the same context, and thus provide no benefit from additional samples.

\section{Unsuccessful Variants}
We also explored several alternative training strategies and architectural variants that did not improve performance; we report them briefly for completeness.

\paragraph{Reduced noise rank.}
We assumed that the input noise tensor of shape $[D, T, H/P, W/P]$ provides maximal stochasticity, and that the dataset's stochasticity might be lower. We therefore sampled noise only for the first frame, $[D, 1, H/P, W/P]$, and duplicated it across time. This reduced-noise variant consistently underperformed and was dropped.

\paragraph{$\ell_1$ vs.\ $\ell_2$ loss.}
We replaced the standard $\ell_2$ flow-matching objective with an $\ell_1$ loss. The motivation was that $\ell_2$ heavily penalizes large deviations, while our predictions are often slightly off the data manifold; we hypothesized the constant-magnitude gradients of $\ell_1$ might stabilize training. In practice, the change degraded performance.

\section{Temporal Consistency Loss}

In video prediction we require \emph{temporal coherence} across the entire sequence.  The goal is not only to predict each frame correctly, but to ensure that the \emph{predicted temporal dynamics} match those of real video latents.

Formally, let the target latent sequence be
\[
x_1 = (x_1^{(1)}, x_1^{(2)}, \ldots, x_1^{(T)}),
\]
with $x_1^{(t)}$ the latent of frame $t$.

Define the temporal finite-difference operator
\[
\Delta x_1^{(t)} = x_1^{(t+1)} - x_1^{(t)}.
\]

A temporally consistent predictor should satisfy
\[
\Delta \hat{x}_1^{(t)} \approx \Delta x_1^{(t)}, \qquad \forall\, t = 1,\dots,T-1,
\]
where $\hat{x}_1$ is the output of the flow-matching sampler.

In other words, beyond matching the final latent $x_1$, the model should reproduce the \emph{temporal derivatives} (or motion patterns) implicit in the data:
\[
\frac{d}{dt}\hat{x}_1(t) \approx \frac{d}{dt} x_1(t).
\]

This encourages:
\begin{itemize}
    \item consistent object motion across time,
    \item reduced flickering and temporal artifacts,
    \item dynamics that align with the true evolution of the latent sequence.
\end{itemize}

Such temporal regularisation can be added as an auxiliary loss or incorporated directly into the velocity field parameterisation to enforce dynamical consistency.

\section{Task specific Loss}
\label{sec:appendix_task_driven_loss}

\paragraph{Time-Dependent Detector Weighting.}
During sampling, early steps determine coarse structure (which objects appear and where), while later steps refine higher-frequency details. Backpropagating the detector loss too early in the sampling process can overconstrain the model. A natural solution is to weight the detector loss with a factor $\lambda_\tau$ based on the timestep $\tau$. We modify the training objective to:
\[
\mathcal{L} = \mathcal{L}_{\text{flow}} + \lambda_\tau\, \mathcal{L}_{\text{det}},
\]
and tune the schedule $\lambda_\tau$.

\begin{figure}[H]
  \vskip 0.2in
  \begin{center}
    \includegraphics[width=0.35\linewidth]{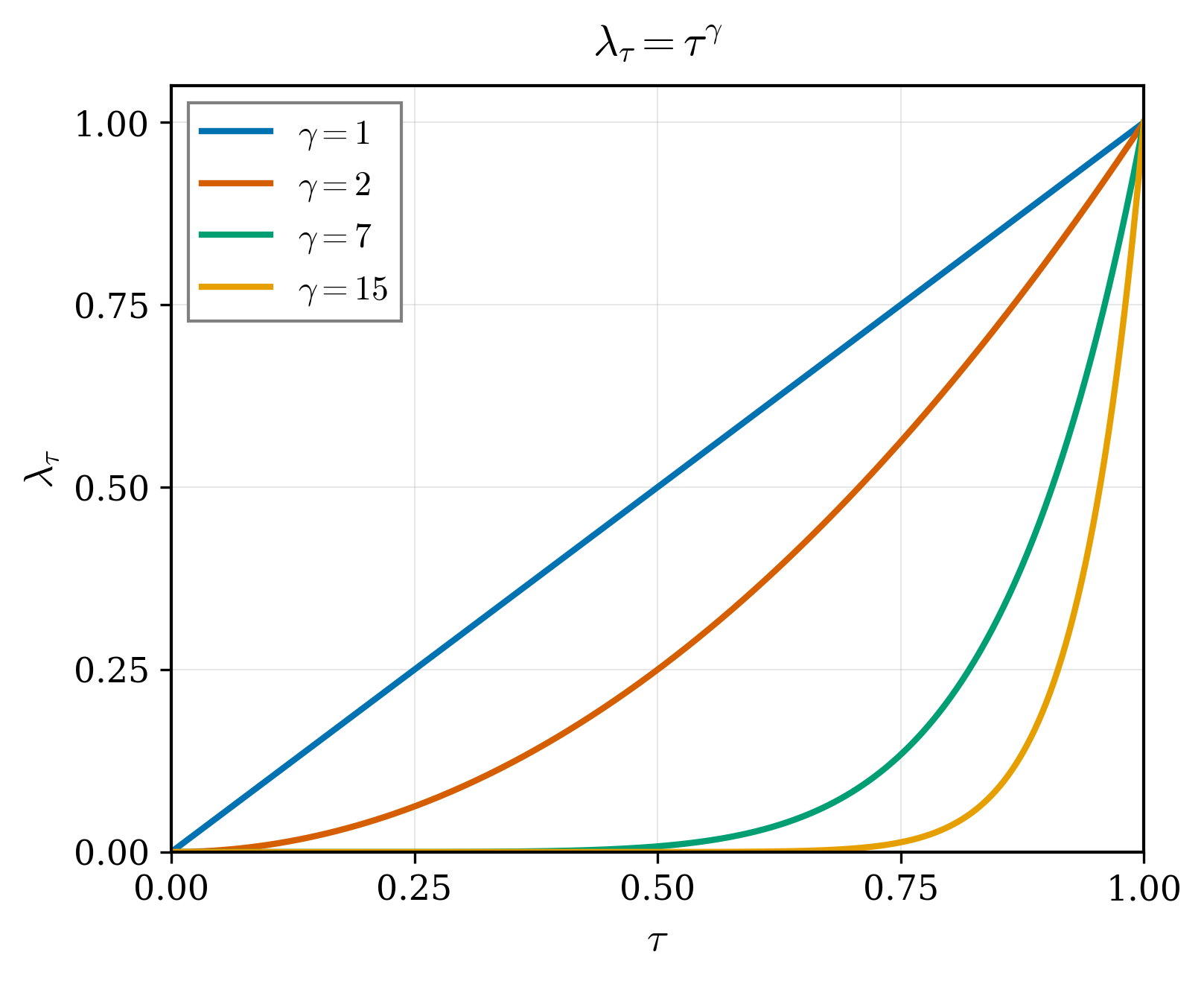}
    \caption{$\lambda_\tau$ for different values of $\gamma$. Larger $\gamma$ increasingly emphasizes late timesteps.}
    \label{fig:gamma_functions}
  \end{center}
\end{figure}

\paragraph{Timestep weighting schedule for detector loss.}
We ablate the choice of the timestep-dependent weight $\lambda_\tau$ used for task-specialized training with the detector loss. We consider schedules of the form $\lambda(\tau) = \tau^\gamma$, illustrated in Figure~\ref{fig:gamma_functions}. Larger values of $\gamma$ place more weight on the detector loss at late timesteps (low-noise regime), where predicted latents are closer to the final semantic representation required by the detector.

\section{One Step Projection}

Using $x_\tau=(1-\tau)x_0+\tau x_1$, we obtain the key identity
\begin{align}
x_{1,\mathrm{pred}} - x_1
&= x_\tau - x_1 + (1-\tau)u_\theta(x_\tau,\tau) \nonumber\\
&= (1-\tau)\big(u_\theta(x_\tau,\tau) - (x_1-x_0)\big).
\label{eq:proj_error_identity}
\end{align}
In particular, for squared error,
\begin{equation}
\|x_{1,\mathrm{pred}}-x_1\|^2
= (1-\tau)^2 \,\big\|u_\theta(x_\tau,\tau) - (x_1-x_0)\big\|^2,
\label{eq:proj_equiv_weighted_fm}
\end{equation}
i.e., an endpoint loss implicitly induces a time-dependent weighting.

\paragraph{Auxiliary losses and time scaling.}
Let $R(\cdot)$ be any differentiable auxiliary loss applied at the predicted endpoint, e.g.,
\[
\mathcal{L}_{\mathrm{aux}}(\tau) = R(x_{1,\mathrm{pred}}).
\]
Using the one-step projection
\[
x_{1,\mathrm{pred}} = x_\tau + (1-\tau)\,u_\theta(x_\tau,\tau),
\]
the chain rule yields a one-step parameter gradient (no unrolling):
\begin{equation}
\nabla_\theta \mathcal{L}_{\mathrm{aux}}
=
(1-\tau)\left(\frac{\partial u_\theta(x_\tau,\tau)}{\partial \theta}\right)^{\!\top}
\nabla_{x_{1,\mathrm{pred}}} R.
\label{eq:aux_grad_one_step}
\end{equation}
Therefore, auxiliary losses applied to $x_{1,\mathrm{pred}}$ are naturally down-weighted near $\tau\to 1$ by a factor $(1-\tau)$ (and often $(1-\tau)^2$ for losses locally proportional to the endpoint error). In practice, this motivates compensating strategies such as modifying the time-sampling distribution or explicitly reweighting $\mathcal{L}_{\mathrm{aux}}(\tau)$ as a function of $\tau$.

\section{Backpropagation through time}

\subsection{Backpropagation through time}
Sampling from a learned flow is typically performed by numerically integrating the ODE
\[
\frac{d x_\tau}{d\tau} = u_\theta(x_\tau,\tau),
\]
which, under an Euler discretization with time points $\{\tau_k\}_{k=0}^{N}$ and $\Delta \tau_k = \tau_{k+1}-\tau_k$, reads
\begin{equation}
x_{k+1} = x_k + \Delta \tau_k\, u_\theta(x_k,\tau_k), \qquad k=0,\dots,N-1.
\label{eq:euler_update}
\end{equation}
If an auxiliary objective $\mathcal{L}$ is applied to the final state (e.g., $\mathcal{L}=\ell(x_N)$), then the gradient $\nabla_\theta \mathcal{L}$ can be computed by backpropagating through the entire integration trajectory (BPTT). Define the adjoint variables
\[
\lambda_k \coloneqq \nabla_{x_k}\mathcal{L}.
\]
Because $x_{k+1}$ depends on $x_k$, the chain rule through one step gives
\begin{equation}
\lambda_k
= \left(\frac{\partial x_{k+1}}{\partial x_k}\right)^{\!\top}\lambda_{k+1}.
\label{eq:bptt_chain_one_step}
\end{equation}
From~\eqref{eq:euler_update}, differentiating w.r.t.\ $x_k$ yields
\[
\frac{\partial x_{k+1}}{\partial x_k}
= I + \Delta \tau_k\, \nabla_x u_\theta(x_k,\tau_k),
\]
hence the BPTT recursion
\begin{equation}
\lambda_k
= \Big(I + \Delta \tau_k\, \nabla_x u_\theta(x_k,\tau_k)\Big)^{\!\top}\lambda_{k+1}.
\label{eq:bptt_core}
\end{equation}
The parameter gradient accumulates contributions from every step where $\theta$ appears:
\begin{equation}
\nabla_\theta \mathcal{L}
= \sum_{k=0}^{N-1} \Delta \tau_k \Big(\nabla_\theta u_\theta(x_k,\tau_k)\Big)^{\!\top}\lambda_{k+1}.
\label{eq:bptt_grad_theta}
\end{equation}

\paragraph{Practical limitations.}
While BPTT provides the exact gradient of the unrolled sampler, it is often difficult to use in practice: (i) it requires storing (or checkpointing) intermediate activations along the trajectory, leading to large memory overhead; (ii) the product of Jacobians in~\eqref{eq:bptt_core} can cause vanishing or exploding gradients; and (iii) it requires running the full sampler inside training. For example, using $N=50$ integration steps makes each training iteration $\mathcal{O}(50)$ times more expensive than standard teacher-forced flow matching.

Previous work \cite{prabhudesai2023aligning} has explored backpropagating through a randomized number of steps K, while using LORA for finetuning less weights.

Reward Feedback Learning (ReFL)~\cite{he2023rewardfeedbacklearning} proposes a more aggressive truncation strategy for diffusion models. Specifically, ReFL samples a random diffusion timestep $t$, generates an on-policy latent $x_t$ by unrolling the diffusion process from pure noise to time $t$ without gradient tracking, and then performs a single additional denoising step with gradients enabled. This design reduces the learning signal to a single-step gradient, significantly lowering memory requirements and avoiding deep Jacobian chains. However, it still incurs the computational cost of running the diffusion sampler up to time $t$ at each iteration, which remains expensive for large numbers of diffusion steps.

\section{Discussion and Limitations}
\label{sec:appendix_limitation}

Our results highlight several open challenges and directions for future work:

\begin{enumerate}
    \item \textbf{Diffusability and structure of latent spaces.}
    Although we demonstrate that Flow Matching can operate directly in high-dimensional semantic latents, it is likely that some representation spaces are inherently more compatible with diffusion or flow-based modeling than others. Understanding which statistical properties make a latent space ``diffusable,'' and how to optimize these properties, potentially through end-to-end training remains an open problem.

    \item \textbf{Toward end-to-end world models.}
    Our current approach relies on a frozen pretrained encoder such as DINOv3. This simplifies training and avoids objective collapse, but also limits the representational quality of the predicted futures to that of the encoder. Recent works such as V-JEPA2 demonstrate that it is possible to train powerful general-purpose encoders, yet their predictors remain deterministic and difficult to scale. Unlocking stable end-to-end training of stochastic world models in high-dimensional semantic spaces represents an important challenge for the field.

    \item \textbf{Action-conditioned world models and planning.}
    The world model developed here is conditioned on a context video. Extending it to action-conditioned settings would enable planning. With stochastic world models now tractable via Flow Matching, a key open question is how to \emph{effectively exploit} this stochasticity: for instance, how to use sampled futures to efficiently explore the space of possible outcomes, generate high-quality trajectory proposals, or guide downstream optimization methods such as Cross-Entropy or model-predictive control.
\end{enumerate}

\section{Broader Impact Statement}
\label{sec:broader_impact}
This paper presents a methodological contribution to latent-space world modeling, focusing on stochastic generative modeling in high-dimensional pretrained semantic representations. The primary goal of this work is to advance the understanding of how uncertainty, multimodality, and semantic structure can be preserved in future prediction models evaluated through downstream perception tasks.

\paragraph{Potential Positive Impacts.}
By improving the ability of world models to represent multiple plausible futures and avoid deterministic mode collapse, this work may benefit research in areas such as robotics, autonomous systems, and model-based reinforcement learning. In particular, more faithful uncertainty and stochasticity modeling may help long horizon planning. Our benchmarks introduced in this work aim to evaluate these capacities.

\paragraph{Potential Risks and Misuse.}
The methods proposed in this paper are general-purpose modeling techniques and do not introduce new sensing, surveillance, or decision-making capabilities. However, as with many advances in predictive modeling, improved future forecasting could be incorporated into safety-critical applications such as autonomous driving or robotics. Misuse, over-reliance on model predictions, or deployment without appropriate safeguards could contribute to unsafe behavior if uncertainty is misinterpreted or ignored. Importantly, this work does not address control, policy learning, or real-time decision-making, and should not be viewed as a complete or deployable autonomous system.

\paragraph{Data, Privacy, and Bias Considerations.}
The experiments rely exclusively on existing datasets and pretrained representations. In particular, our \ourtask{} is derived from the Waymo Open Dataset, which is publicly released and subject to its own data governance, licensing, and privacy protections. This work does not introduce new data collection, does not perform identity recognition, and does not aim to infer sensitive personal attributes. Any biases present in pretrained encoders or datasets may be inherited by the learned world models; mitigating such biases remains an important open research challenge beyond the scope of this paper.

\paragraph{Scope and Limitations.}
This work is intended as a research contribution rather than a deployable system. While stochastic modeling improves mode coverage and robustness over long horizons, it does not guarantee correctness or completeness of future predictions, particularly in highly uncertain or partially observed environments.

Overall, this paper advances foundational research on stochastic world modeling in semantic latent spaces. While the methods may support future applications with societal impact, their ethical implications are consistent with those commonly encountered in advancing general machine learning methodologies, and no specific risks beyond those discussed above require special mitigation at this stage.

\end{document}